\documentclass[11pt,draftcls,onecolumn]{IEEEtran}
\usepackage{amsthm,amssymb,epsfig,color,times}
\usepackage{amsmath}
\usepackage{subfigure}
\usepackage{algorithmic,algorithm}

\def\xbf{\mathbf{x}}

\def\pibf{\boldsymbol{\pi}}

\def\Ibf{\boldsymbol{I}}

\def\Sigmabf{\boldsymbol{\Sigma}}

\def\mbf{\mathbf{m}}
\def\Mbf{\mathbf{M}}

\def\rbf{\mathbf{r}}

\def\Xcal{\mathcal{X}}
\def\Ncal{\mathcal{N}}
\def\obf{\mathbf{o}}
\def\ebf{\mathbf{e}}

\def\Obf{\mathbf{O}}

\def\ubf{\mathbf{u}}

\def\Hcal{\mathcal{H}}
\def\Ubf{\mathbf{U}}
\def\Ibf{\mathbf{I}}

\def\Kbf{\mathbf{K}}
\def\0bf{\mathbf{0}}
\def\1bf{\mathbf{1}}
\def\Real{\mathbb{R}}
\def\Natural{\mathbb{N}}

\DeclareMathOperator{\diag}{diag}
\DeclareMathOperator{\range}{range}

\DeclareMathOperator*{\logdet}{log\,det}

\DeclareMathOperator{\indicator}{I}

\DeclareMathOperator{\pr}{Pr}

\newtheorem{proposition}{Proposition}

\theoremstyle{remark}\newtheorem{remark}{Remark}

\begin{document}
\title{Robust Clustering Using \\ Outlier-Sparsity Regularization}

\author{Pedro A. Forero,~\IEEEmembership{Student Member,~IEEE,} Vassilis Kekatos,~\IEEEmembership{Member,~IEEE,} and\\
Georgios B. Giannakis (Contact Author)*,~\IEEEmembership{Fellow,~IEEE}%
\thanks{Part of this work was presented at the 36th IEEE Intl. Conf. on Acoustics, Speech, and Signal Processing, Prague, Czech Republic, May 2011. Work was in part supported by NSF grant CCF-1016605; Dr. Kekatos was funded by the European Community's Seventh Framework Programme (FP7/2008 under grant agreement No. 234914). The authors are with the ECE Dept., University of Minnesota, Minneapolis, MN 55455, USA, Emails:\{forer002,kekatos,georgios\}@umn.edu}%
}


\maketitle 

\begin{abstract}
Notwithstanding the popularity of conventional clustering algorithms such as K-means and probabilistic clustering, their clustering results are sensitive to the presence of outliers in the data. Even a few outliers can compromise the ability of these algorithms to identify meaningful hidden structures rendering their outcome unreliable. This paper develops robust clustering algorithms that not only aim to cluster the data, but also to identify the outliers. The novel approaches rely on the infrequent presence of outliers in the data which translates to sparsity in a judiciously chosen domain. Capitalizing on the sparsity in the outlier domain, outlier-aware robust K-means and probabilistic clustering approaches are proposed. Their novelty lies on identifying outliers while effecting sparsity in the outlier domain through carefully chosen regularization. A block coordinate descent approach is developed to obtain iterative algorithms with convergence guarantees and small excess computational complexity with respect to their non-robust counterparts. Kernelized versions of the robust clustering algorithms are also developed to efficiently handle high-dimensional data, identify nonlinearly separable clusters, or even cluster objects that are not represented by vectors. Numerical tests on both synthetic and real datasets validate the performance and applicability of the novel algorithms.
\end{abstract}

\vspace*{-1em}


\newpage

\section{Introduction}\label{sec:intro}
Clustering aims to partition a set of data into subsets, called clusters, such that data assigned to the same cluster are similar in some sense. Working with unlabeled data and under minimal assumptions makes clustering a challenging, yet universal tool for revealing data structures in a gamut of applications such as DNA microarray analysis and bioinformatics, (social) network analysis, image processing, and data mining \cite{XuWu05}, 
\cite{hastie09}. Moreover, clustering can serve as a pre-processing step for supervised learning algorithms in applications where labeling data one-at-a-time is costly. Multiple interpretations across disciplines of what a cluster is, have led to an abundant literature of application-specific algorithms \cite{XuWu05}. 

Among the algorithms which cluster data represented by vectors, K-means and Gaussian mixture model (GMM-)based clustering are two popular schemes \cite{Ll82}, \cite{XuWu05}. Conventional K-means relies on the Euclidean distance as a similarity measure, thereby yielding partitions that minimize the within-cluster scatter \cite{hastie09}. Contrastingly, soft (a.k.a. fuzzy) K-means is tailored to identify overlapping clusters by allowing each datum to belong to multiple clusters \cite{Be81}.  GMM-based clustering considers observed data drawn from a probability density function (pdf) following a GMM, where each class-conditional pdf corresponds to a cluster \cite{XuWu05}. Clustering arises as a by-product of a maximum likelihood (ML) estimation framework for the GMM parameters. ML parameter estimates are typically obtained through the expectation-maximization (EM) algorithm \cite{Demp77EM}. Kernel methods have been devised to enable clustering of nonlinearly separable clusters \cite{ScSmMu98}, \cite{ScSm02}.    

Notwithstanding their popularity, K-means and GMM-based clustering are sensitive to inconsistent data, termed outliers, due to their functional dependency on the Euclidean distance \cite{huber}. Outliers appear infrequently in the data, emerging either due to reading errors or because they belong to rarely-seen and hence, markedly informative phenomena. However, even a few outliers can render clustering unreliable: cluster centers and model parameter estimates can be severely biased, and thus the data-to-cluster assignment is deteriorated. This motivates robustifying clustering approaches against outliers at affordable computational complexity in order to unravel the underlying structure in the data. 

Robust clustering approaches to clustering have been investigated. In \cite{DaKr97} and  \cite{Honda10robPCAkmeans}, an additional cluster intended for grouping outliers is introduced with is centroid assumed equidistant from all non-outlying data. Possibilistic clustering measures the so-called typicality of each datum with respect to (wrt) each cluster to decide whether a datum is an outlier \cite{KrKe93}, \cite{PaPaKeBe05}. However, possibilistic clustering is sensitive to initialization and can output the same cluster more than once. Clustering approaches originating from robust statistics, such as the minimum volume ellipsoid and Huber's $\epsilon$-contaminated model-based methods \cite{jolion91GMVE}, \cite{Zhuang96epsiloninsensitive}, extract one cluster at a time. This deflation approach can hinder the underlying data structure by removing elements before seeking other clusters. Other approaches rooted on robust statistics are based on the $\ell_1$-distance (K-medians), Tukey's biweighted function, and trimmed means \cite{Bobrowski91L1}, \cite{Kersten99orderstatistics}, \cite{frigui99tuckey}, \cite{YaWuHsYu08}; but are all limited to linearly separable clusters.

The first contribution of the present work is to introduce a data model for clustering that explicitly accounts for outliers via a deterministic outlier vector per datum (Section \ref{sec:problem_formulation}). A datum is deemed an outlier if its corresponding outlier vector is nonzero. Translating the fact that outliers are rare to {\it sparsity} in the outlier vector domain leads to a neat connection between clustering and the compressive sampling (CS) paradigm \cite{CaTa2006}. Building on this model, an outlier-aware clustering methodology is developed for clustering both from the deterministic (K-means), and the probabilistic (GMMs) perspectives.

The second contribution of this work comprises various iterative clustering algorithms developed for robust hard K-means, soft K-means, and GMM-based clustering (Section \ref{sec:solvers}). The algorithms are based on a block coordinate descent (BCD) iteration and yield closed-form updates for each set of optimization variables. In particular, estimating the outliers boils down to solving a group-Lasso problem \cite{YuLi06}, whose solution is computed in closed form. The novel robust clustering algorithms operate at an affordable computational complexity of the same order as the one for their non-robust counterparts.

Several contemporary applications in bioinformatics, (social) network analysis, image processing, and machine learning call for outlier-aware clustering of high-dimensional data, or involve nonlinearly separable clusters. To accommodate these clustering needs, the novel algorithms are kernelized in Section \ref{sec:high_d}; and this is the third contribution of our work. The assumed model not only enables such a kernelization for both K-means and the probabilistic setups, but it also yields iterative algorithms with closed-form updates. In Section \ref{sec:tests}, the algorithms developed are tested using synthetic as well as real datasets from handwritten digit recognition systems and social networks. The results corroborate the effectiveness of the methods. Conclusions are drawn in Section \ref{sec:conclusions}.

\emph{Notation:} Lower-(upper-)case boldface letters are reserved for column vectors (matrices), and calligraphic letters for sets; $(\cdot)^T$ denotes transposition; $\Natural_N$ the set of naturals $\{1,\ldots,N\}$; $\0bf_p$ ($\1bf_p$) the $p\times 1$ vector of all zeros (ones); $\mathbf{I}_p$ the $p\times p$ identity matrix; $\diag(x_1,\ldots,x_p)$ a $p\times p$ diagonal matrix with diagonal entries $x_1,\ldots,x_p$; $\range(\mathbf{X})$ the range space of matrix $\mathbf{X}$; $\mathbb{E}[\cdot]$ denotes the expectation operator; $\mathcal{N}(\xbf;\mathbf{m},\boldsymbol{\Sigma})$ denotes the multivariate Gaussian pdf with mean $\mathbf{m}$ and covariance matrix $\boldsymbol{\Sigma}$ evaluated at $\xbf$; $\|\mathbf{x}\|_\mathbf{A}:=\sqrt{\mathbf{x}^T\mathbf{A}\mathbf{x}}$ for a positive semidefinite matrix $\mathbf{A}$; $\|\mathbf{x}\|_p:=\left(\sum_{i=1}^n |x_i|^p\right)^{1/p}$ for $p\geq 1$ stands for the $\ell_p$-norm in $\mathbb{R}^n$.


\section{Sparsity-Aware Clustering: Context and Criteria}\label{sec:problem_formulation}

After reviewing the clustering task, a model pertinent to outlier-contaminated data is introduced next. Building on this model, robust approaches are developed for K-means (Section \ref{susbec:hard}) as well as probabilistic clustering (Section \ref{subsec:probabilistic}).

\subsection{K-means Clustering}\label{susbec:hard} 
Given a set of $p$-dimensional vectors $\Xcal:=\{\xbf_1,\ldots,\xbf_N\}$, let $\{\mathcal{X}_1,\ldots,\mathcal{X}_C\}$ be a \emph{partition} of $\mathcal{X}$ to $C$ subsets (clusters) $\mathcal{X}_c\subset \mathcal{X}$ for $c\in \Natural_C$, which are collectively exhaustive, mutually exclusive, and non-empty. Partitional clustering seeks a partition of $\mathcal{X}$ such that two vectors assigned to the same cluster are closer to each other in some well-defined sense, such as the Euclidean distance, than to vectors assigned to other clusters.

Among partitional clustering methods, K-means is one of the most widely used with well-documented merits and a long history \cite{bock07kmeans}. In the K-means setup, a centroid $\mbf_c\in\Real^p$ is introduced per cluster $\mathcal{X}_c$. Then, instead of comparing distances between pairs of points in $\mathcal{X}$, the point-centroid distances $\|\mathbf{x}_n-\mathbf{m}_{c}\|_2$ are considered. Moreover, for each input vector $\mathbf{x}_n$, K-means introduces the unknown memberships $u_{nc}$ for $c\in \Natural_C$, defined to be 1 when $\mathbf{x}_n\in\mathcal{X}_c$, and 0 otherwise. To guarantee a valid partition, the membership coefficients apart from being binary \textbf{(c1)}: $u_{nc}\in\{0,1\}$; they should also
satisfy the constraints \textbf{(c2)}: $\sum_{n=1}^Nu_{nc}>0$, for all $c$, to preclude empty clusters; and \textbf{(c3)}: $\sum_{c=1}^C u_{nc}=1$, for all $n$, so that each vector is assigned to a cluster.

The K-means clustering task can be then posed as that of finding the centroids $\{\mathbf{m}_c\}_{c=1}^C$ and the cluster assignments $u_{nc}$'s by solving the optimization problem
\begin{equation}\label{eq:hard}
\min_{\{\mathbf{m}_c\},\{u_{nc}\}} ~ \sum_{n=1}^N
\sum_{c=1}^C u_{nc}\left\|\xbf_n-\mbf_c\right\|_2^2~\textrm{subject to}~ \textrm{(c1)-(c3)}.
\end{equation}
However, problem \eqref{eq:hard} is known to be NP-hard, even for $C=2$ \cite{DaFr09}. Practically, a suboptimal solution is pursued using the celebrated K-means algorithm. This algorithm drops the (c2) constraint, which is checked in a post-processing step instead. Then, it alternately minimizes the cost in \eqref{eq:hard} wrt one of the set of variables $\{\mathbf{m}_c\}$ and $\{u_{nc}\}$, while keeping the other one fixed, and iterates. K-means iterations are guaranteed to converge to a stationary point of \eqref{eq:hard} \cite{selim84kmeansconv}.

To better motivate and further understand the pros and cons of K-means clustering, it is instructive to postulate a pertinent data model. Such a model assumes that the input vectors can be expressed as $\mathbf{x}_n=\sum_{c=1}^C u_{nc}\mathbf{m}_c + \mathbf{v}_n$, where $\mathbf{v}_n$ is a zero-mean vector capturing the deviation of $\mathbf{x}_n$ from its associated centroid $\mathbf{m}_c$.  It is easy to see that under (c1)-(c3), the minimizers of \eqref{eq:hard} offer merely a blind least-squares (LS) fit of the data $\{\xbf_n\}_{n=1}^N$ respecting the cluster assignment constraints. However, such a simplistic, yet widely applicable model, does not take into account \emph{outliers}; that is points $\mathbf{x}_n$ violating the assumed model. This fact paired with the sensitivity of the LS cost to large residuals explain K-means' vulnerability to outliers \cite{DaKr97}.

To robustify K-means, consider the following data model which explicitly accounts for outliers
\begin{equation}\label{eq:model}
\xbf_n=\sum_{c=1}^Cu_{nc}\mbf_c+\mathbf{o}_n+\mathbf{v}_n,~~~n\in \Natural_N
\end{equation}
where the outlier vector $\mathbf{o}_n$ is defined to be deterministically nonzero if
$\mathbf{x}_n$ corresponds to an outlier, and $\mathbf{0}_p$ otherwise. The unknowns $\{u_{nc},\mathbf{m}_c,\mathbf{o}_n\}$ in \eqref{eq:model} can now be estimated using the LS approach as the minimizers of $\sum_{n=1}^N \left\|\xbf_n{-}\sum_{c=1}^C u_{nc}\mbf_c {-}\obf_n\right\|_2^2$, which are the maximum likelihood (ML) estimates if $\mathbf{v}_n\sim \mathcal{N}(\mathbf{0},\sigma^2\mathbf{I}_p)$. Even if $u_{nc}$'s
were known, estimating $\{\mathbf{m}_c\}$ and $\{\mathbf{o}_n\}$ based solely on $\{\mathbf{x}_n\}$ would be an under-determined problem. The key observation here is that most of the $\{\mathbf{o}_n\}$ are zero. This motivates the following criterion for clustering and identification of at most $s\in\Natural_N$ outliers
\begin{equation}\label{eq:problem1}
\min_{\Mbf,\Obf,\Ubf\in\mathcal{U}_1} ~ \sum_{n=1}^N \left\|\xbf_n-\sum_{c=1}^C u_{nc}\mbf_c -\obf_n\right\|_2^2 ~~ \textrm{s. to}~\sum_{n=1}^N \indicator\left(\|\mathbf{o}_n\|_2 > 0\right)\leq s
\end{equation}
where $\mathbf{M}:=[\mathbf{m}_1~\cdots~\mathbf{m}_C]$, $\mathbf{O}:= [\mathbf{o}_1~\cdots~\mathbf{o}_N]$, $\mathbf{U}\in \mathbb{R}^{N\times C}$ denotes the membership
matrix with entries $[\mathbf{U}]_{n,c} :=u_{nc}$, $\mathcal{U}_1$ is the set of all $\mathbf{U}$ matrices satisfying (c1) and (c3), and $\indicator(\cdot)$ denotes the indicator function. Due to (c1) and (c3), each summand in the cost of \eqref{eq:problem1} can be rewritten as $\sum_{c=1}^C u_{nc}\|\mathbf{x}_n-\mathbf{m}_c-\mathbf{o}_n\|_2^2$; and the Lagrangian form of \eqref{eq:problem1} as 
\begin{align}\label{eq:problem2}
\min_{\Mbf,\Obf,\Ubf\in\mathcal{U}_1} ~&~ \sum_{n=1}^N\sum_{c=1}^C
u_{nc}\left\|\xbf_n-\mbf_c-\obf_n\right\|_2^2 + \lambda\sum_{n=1}^N
\indicator\left(\|\mathbf{o}_n\|_2 > 0\right)
\end{align}%
where $\lambda\geq 0$ is an outlier-controlling parameter. For $\lambda=0$, $\mathbf{o}_n$ should be set equal to the generally nonzero value $\mathbf{x}_n-\mathbf{m}_c$ for any $c$ and yield a zero optimum cost, in which case all $\xbf_n$'s are declared as outliers. When $\lambda\rightarrow \infty$, the optimum $\mathbf{o}_n$'s are zero, all the $\mathbf{x}_n$'s are deemed outlier free, and the problem in \eqref{eq:problem2} reduces to the K-means cost in \eqref{eq:hard}. This reduction of the NP-hard K-means problem to an instance of the problem in \eqref{eq:problem2} establishes the NP-hardness of the latter.

Along the lines of K-means, similar iterations could be pursued for suboptimally solving \eqref{eq:problem2}. However, such iterations cannot provide any convergence guarantees due to the discontinuity of the indicator function at zero. Aiming at a practically feasible solver of \eqref{eq:problem2}, consider first that $\mathbf{U}\in\mathcal{U}_1$ is given. The optimization wrt $\{\mathbf{M},\mathbf{O}\}$ remains non-convex due to $\sum_{n=1}^N\indicator(\|\mathbf{o}_n\|_2> 0)$. Following the successful CS paradigm, where the $\ell_0$-(pseudo)norm of a vector $\mathbf{x}\in\mathbb{R}^N$, defined as $\|\mathbf{x}\|_0:=\sum_{n=1}^N\indicator(|x_n|>0)$, was surrogated by its convex $\ell_1$-norm $\|\mathbf{x}\|_1$, the problem in \eqref{eq:problem2} is replaced by 
\begin{equation}\label{eq:robust_hard}
\min_{\mathbf{M},\mathbf{O},\Ubf\in\mathcal{U}_1}~~
\sum_{n=1}^N\sum_{c=1}^C u_{nc}\left\|\xbf_n-\mbf_c-\obf_n\right\|_2^2+\lambda\sum_{n=1}^N\left\|\obf_n\right\|_2.
\end{equation}
Our robust K-means approach is to minimize \eqref{eq:robust_hard}, which is convex in $\{\mathbf{M},\mathbf{O}\}$, but remains jointly non-convex. The algorithm for suboptimally solving the non-convex problem in \eqref{eq:robust_hard} is postponed for Section \ref{subsec:kmeansolver}. Note that the minimization in \eqref{eq:robust_hard} resembles the group Lasso criterion used for recovering a block-sparse vector in a linear regression setup \cite{YuLi06}. This establishes an interesting link between robust clustering and CS. A couple of remarks are now in order.

\begin{remark}[{\it Mahalanobis distance}]
If the covariance matrix of $\mathbf{v}_n$ in \eqref{eq:model} is known, say $\mathbf{\Sigma}$, the Euclidean distance in \eqref{eq:problem1}-\eqref{eq:robust_hard} can be replaced by the
Mahalanobis distance $\|\mathbf{x}_n{-}\mathbf{m}_c{-}\mathbf{o}_n\|_{\mathbf{\Sigma}^{-1}}^2$.
\end{remark}

\begin{remark}[{\it $\ell_1$-penalty for entry-wise outliers}]
The regularization term $\sum_{n=1}^N\|\obf_n\|_2$ in \eqref{eq:robust_hard} enables identifying whole data vectors as outliers. Replacing it by $\sum_{n=1}^N\|\obf_n\|_1$ enables recovery of outlying data entries instead of the whole vector. Iterative solvers for this case can be developed using the methodology presented in Section \ref{sec:solvers}; due to space limitations this case is not pursued here.
\end{remark}

Constraints (c1) and (c3) in \eqref{eq:hard} entail \emph{hard} membership assignments, meaning that each vector is assigned to a single cluster. However, \emph{soft} clustering which allows each vector to partially belong to several clusters, can better identify overlapping clusters \cite{Be81}. One way to obtain fractional memberships is via soft K-means. Soft K-means differs from hard K-means by (i) relaxing the binary-alphabet constraint (c1) to the box constraint {\bf(c4):} $u_{nc}\in[0,1]$; and (ii) by raising the $u_{nc}$'s in \eqref{eq:hard} to the $q$-th power, where $q>1$ is a tuning parameter \cite{Be81}. The robust soft K-means scheme proposed here amounts to replacing $\mathbf{x}_n$ with its outlier-compensated version $(\mathbf{x}_n-\mathbf{o}_n)$, and leveraging the sparsity of the $\{\mathbf{o}_n\}$'s. These steps lead to the following criterion 
\begin{equation}\label{eq:robust_soft}
\min_{\mathbf{M},\mathbf{O},\Ubf\in\mathcal{U}_2} ~~ \sum_{n=1}^N
\sum_{c=1}^C u_{nc}^q\left(\left\|\xbf_n-\mbf_c-\obf_n\right\|_2^2
+\lambda\left\|\obf_n\right\|_2\right)
\end{equation}
where $\mathcal{U}_2$ is the set of all $\mathbf{U}$ matrices satisfying (c3)-(c4). An algorithm for approximately solving \eqref{eq:robust_soft} is presented in Section \ref{subsec:kmeansolver}. Note that a hard partition of $\Xcal$ can still be obtained from the soft $u_{nc}$ by assigning $\xbf_n$ to the $\hat{c}$-th cluster, where $\hat{c}:= \arg\max_{c} u_{nc}$.

\subsection{Probabilistic Clustering}\label{subsec:probabilistic}
An alternative way to perform soft clustering is by following a probabilistic approach \cite{XuWu05}. To this end, a mixture distribution model is postulated for $\mathbf{x}_n$, while $\{u_{nc}\}_{c=1}^C$ are now interpreted as unobserved (latent) random variables. The centroids $\{\mathbf{m}_c\}_{c=1}^C$ are treated as deterministic parameters of the mixture distribution, and their ML estimates are subsequently obtained via the EM algorithm. 

To account for outliers, probabilistic clustering is generalized to model \eqref{eq:model}. Suppose that the $\{\mathbf{x}_n\}$'s in \eqref{eq:model} are i.i.d. drawn from a mixture model where the $\{\mathbf{o}_n\}$'s are deterministic parameters. The memberships $\mathbf{u}_n:=[u_{n1}\cdots u_{nC}]^T$ are latent random vectors, corresponding to the rows of $\Ubf$, and take values in $\{\mathbf{e}_1,\ldots,\mathbf{e}_C\}$, where $\mathbf{e}_c$ is the $c$-th column of $\mathbf{I}_C$. If $\xbf_n$ is drawn from the $c$-th mixture component, then $\mathbf{u}_n=\mathbf{e}_c$. Assume further that the class-conditional pdf's are Gaussian and modeled as $p(\mathbf{x}_n|\mathbf{u}_{n}=\mathbf{e}_c) = \mathcal{N}(\mathbf{x}_n;\mathbf{m}_c{+}\mathbf{o}_n,\mathbf{\Sigma})$ for all $n$ and $c$. This implies that $p(\mathbf{x}_n)= \sum_{c=1}^C\pi_c~ \mathcal{N}\left(\mathbf{x}_n;\mathbf{m}_c{+}\mathbf{o}_n,\mathbf{\Sigma}\right)$ with $\pi_c:=\pr(\mathbf{u}_{n}=\mathbf{e}_c)$. If the $\mathbf{x}_n$'s are independent, the log-likelihood of the input data is
\begin{equation}\label{eq:likelihood}
L(\mathbf{X};\boldsymbol{\pi},\mathbf{M},\mathbf{O},\mathbf{\Sigma}):=\sum_{n=1}^N\log
\left( \sum_{c=1}^C
\pi_c~\mathcal{N}\left(\mathbf{x}_n;\mathbf{m}_c{+}\mathbf{o}_n,\mathbf{\Sigma}\right)
\right)
\end{equation}
where $\mathbf{X}:=\left[\mathbf{x}_1~\cdots~\mathbf{x}_N\right]$, and $\boldsymbol{\pi}:=[\pi_1\cdots\pi_C]^T$. Controlling the number of outliers (number of zero $\mathbf{o}_n$ vectors) suggests minimizing the \emph{regularized} negative log-likelihood as
\begin{equation}\label{eq:robust_MLE}
\min_{\mathbf{\Theta}}
-L\left(\mathbf{X};\mathbf{\Theta}\right)
+\lambda \sum_{n=1}^N\left\|\mathbf{o}_n\right\|_{\mathbf{\Sigma}^{-1}}
\end{equation}
where $\mathbf{\Theta}:= \left\{\boldsymbol{\pi}\in\mathcal{P}, \mathbf{M},\mathbf{O},\mathbf{\Sigma}\succ 0\right\}$ is the set of all model parameters, $\mathcal{P}$ is the probability simplex $\mathcal{P}:= \{\boldsymbol{\pi}:\boldsymbol{\pi}^T\mathbf{1}{=}1 \textrm{ and } \boldsymbol{\pi}\geq \mathbf{0}\}$, and $\mathbf{\Sigma}\succ 0$ means that $\mathbf{\Sigma}$ is a positive definite matrix. An EM-based algorithm for solving \eqref{eq:robust_MLE} is derived in Section \ref{subsec:EMsolver}. Having estimated the parameters of the likelihood, the posterior probabilities $\gamma_{nc}:=\pr(\mathbf{u}_{n}=\mathbf{e}_c|\mathbf{x}_n)$ can be readily obtained and interpreted as soft memberships. 

Although modeling all class conditionals having a common covariance matrix $\mathbf{\Sigma}$ may seem restrictive, it guarantees that the GMM is well-posed, thereby avoiding spurious unbounded likelihood values \cite[p.~433]{Bi06}. Specifically, it is easy to see that even if all $\obf_n$'s are set to zero, the log-likelihood of a GMM with different covariance matrices $\mathbf{\Sigma}_c$ per mixture grows unbounded, e.g., by setting one of the $\mathbf{m}_c$'s equal to an $\xbf_n$ and letting $\mathbf{\Sigma}_c\rightarrow \mathbf{0}$ for that particular $c$. This possibility for unboundedness is also present in \eqref{eq:robust_MLE}, and justifies the use of a common $\mathbf{\Sigma}$. But even with a common covariance matrix, the vectors $\obf_n$ can drive the log-likelihood in \eqref{eq:likelihood} to infinity. Consider for example, any $(\mbf_c,\mathbf{o}_n)$ pair satisfying  $\mathbf{x}_n=\mbf_c+\mathbf{o}_n$ and let $\mathbf{\Sigma}\rightarrow \mathbf{0}$. To make the problem of maximizing $L\left(\mathbf{X};\mathbf{\Theta}\right)$ well-posed, the $\left\|\mathbf{o}_n \right\|_{\mathbf{\Sigma}^{-1}}$ regularizer is introduced. Note also that for $\lambda\rightarrow \infty$, the optimal $\mathbf{O}$ is zero and \eqref{eq:robust_MLE} reduces to the conventional MLE estimation of a GMM; whereas for $\lambda\rightarrow 0$, the cost in \eqref{eq:robust_MLE} becomes unbounded from below.

\section{Robust Clustering Algorithms}\label{sec:solvers}
Algorithms for solving the problems formulated in Section \ref{sec:problem_formulation} are developed here. Section \ref{subsec:kmeansolver} focuses on the minimization of \eqref{eq:robust_soft}, while the minimization in \eqref{eq:robust_hard} is obtained from \eqref{eq:robust_soft} for $q=1$. In Section \ref{subsec:EMsolver}, an algorithm for minimizing \eqref{eq:robust_MLE} is derived based on the EM approach. Finally, modified versions of the new algorithms with enhanced resilience to outliers are pursued in Section \ref{subsec:weighed}.

\subsection{Robust (Soft) K-Means Algorithms}\label{subsec:kmeansolver}
Consider first solving \eqref{eq:robust_soft} for $q>1$. Although the cost in \eqref{eq:robust_soft} is jointly nonconvex, it is convex wrt each of $\Mbf$, $\Obf$, and $\Ubf$. To develop a suboptimum yet practical solver, the aforementioned per-variable convexity prompted us to devise a BCD algorithm, which minimizes the cost iteratively wrt each optimization variable while holding the other two variables fixed. Let $\Mbf^{(t)}$, $\Obf^{(t)}$, and $\Ubf^{(t)}$ denote the tentative solutions found at the $t$-th iteration. Also, initialize $\mathbf{U}^{(0)}$ randomly in $\mathcal{U}_2$, and $\mathbf{O}^{(0)}$ to zero.

In the first step of the $t$-th iteration, \eqref{eq:robust_soft} is optimized wrt $\Mbf$ for $\Ubf=\Ubf^{(t-1)}$ and $\Obf=\Obf^{(t-1)}$. The optimization decouples over the $\mbf_c$'s, and every $\mbf_c^{(t)}$ is the closed-form solution of an LS problem as
\begin{equation}\label{eq:m_update}
\mbf_c^{(t)}=\frac{\displaystyle\sum_{n=1}^N(u_{nc}^{(t-1)})^q \left(\xbf_n-\obf_n^{(t-1)}\right)}{\displaystyle\sum_{n=1}^N(u_{nc}^{(t-1)})^q}.
\end{equation}

In the second step, the task is to minimize \eqref{eq:robust_soft} wrt $\Obf$ for $\Ubf=\Ubf^{(t-1)}$ and $\Mbf=\Mbf^{(t)}$. The optimization problem decouples per index $n$, so that each $\obf_n$ can be found as the minimizer of
\begin{equation}\label{eq:phi}
\phi^{(t)}(\obf_n):=\sum_{c=1}^C (u_{nc}^{(t-1)})^q \left(\|\xbf_n-\mbf_c^{(t)}
-\obf_n\|_2^2+\lambda\|\obf_n\|_2\right).
\end{equation}
The cost $\phi^{(t)}(\obf_n)$ is convex but non-differentiable. However, its minimizer can be expressed in closed form as shown in the ensuing proposition.
\begin{proposition}\label{pr:phi_minimizer}
The optimization problem in \eqref{eq:phi} is uniquely minimized by
\begin{equation}\label{eq:o_update}
\obf_n^{(t)}=\rbf_n^{(t)}\left[1-\frac{\lambda}{2\|\rbf_n^{(t)}\|_2}\right]_+
\end{equation}
where $[x]_+:=\max\{x,0\}$, and $\rbf_n^{(t)}$ is defined as
\begin{equation}\label{eq:r_update}
\rbf_{n}^{(t)}:=\frac{\sum_{c=1}^C(u_{nc}^{(t-1)})^q\left(\xbf_n-\mbf_c^{(t)}\right)}{\sum_{c=1}^C(u_{nc}^{(t-1)})^q}.
\end{equation}
\end{proposition}
\begin{IEEEproof}
Since $\sum_{c=1}^C (u_{nc}^{(t)})^q>0$ for all $n$ and $t$ due to (c3), the first summand of $\phi^{(t)}(\obf_n)$ in \eqref{eq:phi} is a strictly convex function of $\mathbf{o}_n$. Hence, $\phi^{(t)}(\obf_n)$ is a strictly convex function too and its minimizer is unique. Then, recall that a vector $\mathbf{o}_n^{(t)}$ is a minimizer of \eqref{eq:phi} if and only if $\mathbf{0}\in \partial \phi^{(t)}(\obf_n^{(t)})$, where $\partial \phi^{(t)}(\obf_n)$ is the sub-differential of $\phi^{(t)}(\obf_n)$. For $\mathbf{o}_n\neq \mathbf{0}$, where the cost in \eqref{eq:phi} is differentiable, $\partial\phi^{(t)} (\obf_n)$ is simply the gradient $-2\sum_{c=1}^C (u_{nc}^{(t-1)})^q \left(\mathbf{x}_n-\mathbf{m}_c-\left(1+\frac{\lambda}{2\|\mathbf{o}_n\|_2}\right)\mathbf{o}_n\right)$. At $\mathbf{o}_n=\mathbf{0}$, the sub-differential of the $\ell_2$-norm $\|\mathbf{o}_n\|_2$ is the set of vectors $\{\mathbf{v}_n:\|\mathbf{v}_n\|_2\leq 1\}$ by definition, and then the sub-differential of $\phi^{(t)}(\obf_n)$ is $\partial \phi^{(t)}(\obf_n)=\left\{-2\sum_{c=1}^C (u_{nc}^{(t-1)})^q \left(\mathbf{x}_n-\mathbf{m}_c-\frac{\lambda}{2}\mathbf{v}_n\right):~ \|\mathbf{v}_n\|_2\leq 1\right\}$.

When the minimizer $\mathbf{o}_n^{(t)}$ is nonzero, the condition $\mathbf{0}\in \partial \phi^{(t)}(\obf_n^{(t)})$ implies 
\begin{equation}\label{eq:on_formula}
\left(1+\frac{\lambda}{2\|\mathbf{o}_n^{(t)}\|_2}\right) \mathbf{o}_n^{(t)}=\mathbf{r}_n^{(t)}
\end{equation}
where $\mathbf{r}_n^{(t)}$ has been defined in \eqref{eq:r_update}. Equation \eqref{eq:on_formula} reveals that $\mathbf{o}_n^{(t)}$ is a positively scaled version of $\rbf_{n}^{(t)}$. The scaling can be readily found by taking the $\ell_2$-norm on both sides of \eqref{eq:on_formula}, i.e., $\|\mathbf{o}_n^{(t)}\|_2=\|\mathbf{r}_n^{(t)}\|_2-\lambda/2$, which is valid for $\|\mathbf{r}_n^{(t)}\|_2 > \lambda/2$. Substituting this back to \eqref{eq:on_formula}, yields $\mathbf{o}_n^{(t)}=\mathbf{r}_n^{(t)} \left(1-\frac{\lambda}{2\|\mathbf{r}_n^{(t)}\|_2}\right)$.

For $\mathbf{o}_n^{(t)}=\mathbf{0}$, there exists a $\mathbf{v}_n^{(t)}$ for which $\|\mathbf{v}_n^{(t)}\|_2\leq 1$ and $\mathbf{v}_n^{(t)}=(2/\lambda)\mathbf{r}_n^{(t)}$. This is possible when $\|\mathbf{r}_n^{(t)}\|_2\leq \lambda/2$. These two cases for the minimizer of \eqref{eq:phi} are compactly expressed via \eqref{eq:o_update}.
\end{IEEEproof}

The update for $\mathbf{o}_n^{(t)}$ in \eqref{eq:o_update} reveals two interesting points: (i) the cost $\phi^{(t)}(\obf_n)$ indeed favors zero minimizers; and (ii) the number of outliers is controlled by $\lambda$. After updating vector $\rbf_n^{(t)}$, its  norm is compared against the threshold $\lambda/2$. If $\|\rbf_n^{(t)}\|_2$ exceeds $\lambda/2$, vector $\xbf_n$ is deemed an outlier, and it is compensated by a nonzero $\obf_n^{(t)}$. Otherwise, $\obf_n^{(t)}$ is set to zero and $\xbf_n$ is clustered as a regular point.

During the last step of the $t$-th iteration, \eqref{eq:robust_soft} is minimized over $\Ubf\in\mathcal{U}_2$ for $\Mbf=\mathbf{M}^{(t)}$ and $\Obf=\mathbf{O}^{(t)}$. Similar to the conventional soft K-means, the minimizer is available in closed form as \cite{Be81}
\begin{equation}\label{eq:u_update}
u_{nc}^{(t)}=\left[\sum_{c'=1}^C\left(\frac{\|\xbf_n-\mbf_c^{(t)}-\obf_n^{(t)}\|_2^2+\lambda\|\obf_n^{(t)}\|_2}{\|\xbf_n-\mbf_{c'}^{(t)}-\obf_n^{(t)}\|_2^2+\lambda\|\obf_n^{(t)}\|_2}\right)^{\frac{1}{q-1}}\right]^{-1}.
\end{equation}

Regarding the robust hard K-means, a similar BCD approach for solving \eqref{eq:robust_hard} leads to updating $\Mbf^{(t)}$ and $\Obf^{(t)}$ via \eqref{eq:m_update}, and \eqref{eq:o_update}-\eqref{eq:r_update} for $q=1$. Updating $\Ubf^{(t)}$ boils down to the minimum-distance rule
\begin{equation}\label{eq:u_update_hard}
u_{nc}^{(t)}=\left\{\begin{array}{ll}
1&,~c=\mathop{\arg\min}_{c'}\|\xbf_n-\mbf_{c'}^{(t)}-\obf_n^{(t)}\|_2\\
0&,~\textrm{otherwise}
\end{array}\right..
\end{equation}
Note that \eqref{eq:u_update_hard} is the limit case of \eqref{eq:u_update} for $q\rightarrow 1^+$.

\begin{algorithm}
\caption{Robust K-means} \label{alg:RKM}
\begin{algorithmic}[1]
\REQUIRE Input data matrix $\mathbf{X}$, number of clusters $C$, $q\geq1$, and $\lambda>0$.
\STATE Initialize $\mathbf{O}^{(0)}$ to zero and $\mathbf{U}^{(0)}$ randomly in $\mathcal{U}_2$.
\FOR{$t=1,2,\ldots$}
\STATE Update $\mathbf{M}^{(t)}$ via \eqref{eq:m_update}. 
\STATE Update $\mathbf{O}^{(t)}$ via \eqref{eq:o_update}-\eqref{eq:r_update}. 
\STATE Update $\mathbf{U}^{(t)}$ via \eqref{eq:u_update} ($q>1$) or \eqref{eq:u_update_hard} ($q=1$). 
\ENDFOR
\end{algorithmic}
\end{algorithm}

The robust K-means (RKM) algorithm is tabulated as Algorithm \ref{alg:RKM}. RKM is terminated when $\|\Mbf^{(t)}-\Mbf^{(t-1)}\|_{F}/\|\Mbf^{(t)}\|_{F}\leq\epsilon_s$, where $\|\cdot\|_F$ denotes the Frobenius norm of a matrix, and $\epsilon_s$ is a small positive threshold, e.g., $\epsilon_s=10^{-6}$. The computational resources needed by RKM are summarized next.

\begin{remark}[\it{Computational complexity of RKM}]\label{re:RKM_computations}
Suppose for concreteness that: \textbf{(as1)} the number of clusters is small, e.g., $C<p$; and \textbf{(as2)} the number of points is much larger than the input dimension, i.e., $N\gg p$. When (as2) does not hold, a modification of RKM is developed in Section \ref{sec:high_d}. Under (as1)-(as2), the conventional K-means algorithm performs $\mathcal{O}(NCp)$ scalar operations per iteration, and requires storing $\mathcal{O}(Np)$ scalar variables. For RKM, careful counting shows that the per iteration time-complexity is maintained at $\mathcal{O}(NCp)$: \eqref{eq:u_update} requires computing the $NC$ Euclidean distances $\|\xbf_n-\mbf_c^{(t-1)}-\obf_n^{(t-1)}\|_2^2$ and the $N$ norms $\|\obf_n^{(t-1)}\|_2$ which is $\mathcal{O}(NCp)$; $\mbf_c^{(t)}$'s are updated in $\mathcal{O}(NCp)$; while \eqref{eq:o_update}-\eqref{eq:r_update} entail $\mathcal{O}(NCp)$ operations. Further, the memory requirements of RKM are of the same order as those for K-means. Note also that the additional $N\times p$ matrix $\mathbf{O}$ can be stored using sparse structures.
\end{remark}

The RKM iterations are convergent under mild conditions. This follows because the sequence of cost function values is non-increasing. Since the cost is bounded below, the function value sequences are guaranteed to converge. Convergence of the RKM iterates is characterized in the following proposition.

\begin{proposition}\label{pr:RKMconv}
The RKM algorithm for $q \geq 1$ converges to a coordinate-wise minimum of \eqref{eq:robust_soft}. Moreover, the hard RKM algorithm ($q=1$) converges to a local minimum of \eqref{eq:robust_hard}.
\end{proposition}
\begin{IEEEproof}
By defining $f_s(c)$ as being zero when the Boolean argument $c$ is true, and $\infty$ otherwise, the problem in \eqref{eq:robust_soft} can be written in the unconstrained form
\begin{equation}\label{eq:equiv1}
\min_{\Mbf,\Obf,\mathbf{U}}\sum_{n=1}^N\sum_{c=1}^Cu_{nc}^q\left(\|\xbf_n-\mbf_c-\obf_n\|_2^2+\lambda\|\obf_n\|_2\right)+f_s(\Ubf \in \mathcal{U}_2).
\end{equation}
The cost in \eqref{eq:equiv1}, call it $f(\Mbf,\Obf,\mathbf{U})$, is a proper and lower semi-continuous function, which implies that its non-empty level sets are closed. Also, since $f$ is coercive, its level sets are bounded. Hence, the non-empty level sets of $f$ are compact. For $q > 1$, function $f(\Mbf,\Obf,\Ubf)$ has a unique minimizer per optimization block variable $\Mbf$, $\Obf$, and $\Ubf$. Then, convergence of the RKM algorithm to a coordinate-wise minimum point of \eqref{eq:robust_soft} follows from \cite[Th.~4.1(c)]{Ts01}.

When $q=1$, define the first summand in \eqref{eq:equiv1} as $f_0(\Mbf,\Obf,\Ubf):=\sum_{n=1}^N\sum_{c=1}^Cu_{nc}\|\xbf_n-\mbf_c-\obf_n\|_2^2$, which is the differentiable part of $f$. Function $f_0$ has an open domain, and the remaining non-differentiable part of $f$ is separable wrt the optimization blocks. Hence, again by \cite[Th.~4.1(c)]{Ts01}, the RKM algorithm with $q=1$ converges to a local minimum $(\Mbf^*,\Obf^*,\Ubf^*)$ of \eqref{eq:robust_soft}.

It has been shown so far that for $q=1$, a BCD iteration converges to a local minimum of \eqref{eq:robust_soft}. The BCD step for updating $\Ubf$ is the hard rule in \eqref{eq:u_update_hard}. Hence, this BCD algorithm (i) yields a $\Ubf^*$ with binary entries, and (ii) essentially implements the BCD updates for solving \eqref{eq:robust_hard}. Since a local minimum of \eqref{eq:robust_soft} with binary assignments is also a local minimum of \eqref{eq:robust_hard}, the claim of the proposition follows.
\end{IEEEproof}

\subsection{Robust Probabilistic Clustering Algorithm}\label{subsec:EMsolver}
An EM approach is developed in this subsection to carry out the minimization in \eqref{eq:robust_MLE}. If $\mathbf{U}$ were known, the model parameters $\mathbf{\Theta}$ could be estimated by minimizing the regularized negative log-likelihood of the \emph{complete data} $(\mathbf{X},\mathbf{U})$; that is, 
\begin{equation}\label{eq:EM1}
\min_{\mathbf{\Theta}} ~ - L\left(\mathbf{X},\mathbf{U};\mathbf{\Theta}\right) +\lambda \sum_{n=1}^N \|\mathbf{o}_n\|_{\Sigmabf^{-1}}
\end{equation}
where
\begin{equation}\label{eq:EM2}
L\left(\mathbf{X},\mathbf{U};\mathbf{\Theta}\right):=\sum_{n=1}^N\sum_{c=1}^C u_{nc}\left(\log\pi_c + \log\mathcal{N}\left(\xbf_n;\mbf_c+\obf_n,\Sigmabf\right)\right).
\end{equation}
But since $\mathbf{U}$ is not observed, the cost in \eqref{eq:EM1} is suboptimally minimized by iterating the two steps of the EM method. Let $\mathbf{\Theta}^{(t)}$ denote the model parameter values at the $t$-th iteration. During the E-step of the $t$-th iteration, the expectation $Q(\mathbf{\Theta};\mathbf{\Theta}^{(t-1)}):=\mathbb{E}_{\mathbf{U}|\mathbf{X};\mathbf{\Theta}^{(t-1)}}\left[L\left(\mathbf{X},\mathbf{U};\mathbf{\Theta}\right)\right]$ is evaluated. Since $L\left(\mathbf{X},\mathbf{U};\mathbf{\Theta}\right)$ is a linear function of $\mathbf{U}$, and $u_{nc}$'s are binary random variables, it follows that
\begin{equation}\label{eq:Qfunction} Q(\mathbf{\Theta};\mathbf{\Theta}^{(t-1)})=\sum_{n=1}^N\sum_{c=1}^C\gamma_{nc}^{(t)} \left(\log\pi_c + \log\mathcal{N}\left(\xbf_n;\mbf_c+\obf_n,\Sigmabf\right)\right)
\end{equation}
where $\gamma_{nc}^{(t)}:=\pr\left(\ubf_{n}=\ebf_c|\mathbf{x}_n;\mathbf{\Theta}^{(t-1)}\right)$. Using Bayes' rule, the posterior probabilities $\gamma_{nc}^{(t)}$ are evaluated in closed form as
\begin{equation}\label{eq:gamma_update}
\gamma_{nc}^{(t)}=\frac{\pi_c^{(t-1)}\Ncal(\xbf_n;\mbf_c^{(t-1)}+\obf_n^{(t-1)},\Sigmabf^{(t-1)})}{\sum_{c'=1}^C\pi_{c'}^{(t-1)}\Ncal(\xbf_n;\mbf_{c'}^{(t-1)}+\obf_n^{(t-1)},\Sigmabf^{(t-1)})}.
\end{equation}

During the M-step, $\boldsymbol{\Theta}^{(t)}$ is updated as
\begin{equation}\label{eq:EM_cost}
\mathbf{\Theta}^{(t)}=\mathop{\arg\min}_{\mathbf{\Theta}} ~ - Q(\mathbf{\Theta};\mathbf{\Theta}^{(t-1)}) + \lambda \sum_{n=1}^N\|\mathbf{o}_n\|_{\Sigmabf^{-1}}.
\end{equation}
A BCD strategy that updates each set of the parameters in $\mathbf{\Theta}$ one at a time with all other ones fixed, is described next. First, the cost in \eqref{eq:EM_cost} is minimized wrt $\pibf$. Given that $\sum_{c=1}^C \gamma_{nc}^{(t)}=1$ for all $n$, it is easy to check that the minimizer of $-\sum_{n=1}^N\sum_{c=1}^C\gamma_{nc}^{(t)}\log\pi_c$ over $\mathcal{P}$ is found in closed form as
\begin{equation}\label{eq:pi_update}
\pi_c^{(t)}=\frac{1}{N}\sum_{n=1}^N\gamma_{nc}^{(t)}\textrm{~~for all~~}c\in \Natural_C.
\end{equation}

Subsequently, \eqref{eq:EM_cost} is minimized wrt $\Mbf$ while $\pibf$, $\Obf$, and $\Sigmabf$ are set respectively to $\pibf^{(t)}$, $\Obf^{(t-1)}$, and $\Sigmabf^{(t-1)}$. The centroids are updated as the minimizers of a weighted LS cost yielding
\begin{equation}\label{eq:m_updateEM}
\mbf_c^{(t)}=\frac{\sum_{n=1}^N\gamma_{nc}^{(t)}\left(\xbf_n-\obf_n^{(t-1)}\right)}{\sum_{n=1}^N\gamma_{nc}^{(t)}}\textrm{~~for all~~}c\in \Natural_C.
\end{equation}

Then, \eqref{eq:EM_cost} is minimized wrt $\Obf$ while keeping the rest of the model parameters fixed to their already updated values. This optimization decouples over $n$, and one has to solve
\begin{equation}\label{eq:o_minEM}
\min_{\obf_n}\sum_{c=1}^C\frac{\gamma_{nc}^{(t)}}{2}\|\xbf_n-\mbf_c^{(t)}-\obf_n\|_{(\Sigmabf^{(t-1)})^{-1}}^2 + \lambda\|\obf_n\|_{(\Sigmabf^{(t-1)})^{-1}}
\end{equation}
for all $n\in\Natural_N$. For a full covariance $\Sigmabf$, \eqref{eq:o_minEM} can be solved as a second-order cone program. For the case of \emph{spherical} clusters, i.e., $\Sigmabf=\sigma^2\Ibf_p$, solving \eqref{eq:o_minEM} simplifies considerably. Specifically, the cost can then be written as $\sum_{c=1}^C\gamma_{nc}^{(t)} \|\xbf_n-\mbf_c^{(t)}-\obf_n\|_2^2 + 2\lambda\sigma^{(t-1)}\|\obf_n\|_2$, which is similar to the cost in \eqref{eq:phi} for $q=1$, and for an appropriately scaled $\lambda$. Building on the solution of \eqref{eq:phi}, the $\mathbf{o}_n$'s are updated as
\begin{equation}\label{eq:o_updateEM}
\obf_n^{(t)}=\mathbf{r}_n^{(t)}\left[1-\frac{\lambda\sigma^{(t-1)}}{\|\mathbf{r}_n^{(t)}\|_2}\right]_+
\end{equation}
after redefining the residual vector as $\mathbf{r}_n^{(t)}:= \sum_{c=1}^C\gamma_{nc}^{(t)}(\xbf_n-\mbf_c^{(t)})$ in lieu of \eqref{eq:r_update}. Interestingly, the thresholding rule of \eqref{eq:o_updateEM} shows that $\sigma^{(t-1)}$ affects the detection of outliers. In fact, in this probabilistic setting, the threshold for outlier identification is proportional to the value of the outlier-compensated standard deviation estimate and, hence, it is adapted to the empirical distribution of the data.

The M-step is concluded by minimizing \eqref{eq:EM_cost} wrt $\Sigmabf$ for $\pibf=\pibf^{(t)}$, $\Mbf=\Mbf^{(t)}$, and $\Obf=\Obf^{(t)}$, i.e., 
\begin{equation}\label{eq:Sigma_update}
\min_{\Sigmabf\succ0}\sum_{n=1}^N\sum_{c=1}^C\frac{\gamma_{nc}^{(t)}}{2}\|\xbf_n-\mbf_c^{(t)}-\obf_n^{(t)}\|_{\Sigmabf^{-1}}^2 + \frac{N}{2}\logdet\Sigmabf+\lambda \sum_{n=1}^N\|\obf_n^{(t)}\|_{\Sigmabf^{-1}}.
\end{equation}
For a generic $\Sigmabf$, \eqref{eq:Sigma_update} must be solved numerically, e.g., via gradient descent or interior point methods. Considering \emph{spherical} clusters for simplicity, the first order optimality condition for \eqref{eq:Sigma_update} requires solving a quadratic equation in $\sigma^{(t)}$. Ignoring the negative root of this equation, $\sigma^{(t)}$ is found as
\begin{equation}\label{eq:sigma_update}
\sigma^{(t)}=\frac{\lambda }{2Np}\sum_{n=1}^N\|\obf_n^{(t)}\|_2 + \sqrt{\frac{1}{Np}\sum_{n=1}^N\sum_{c=1}^C\gamma_{nc}^{(t)}\|\xbf_n-\mbf_c^{(t)}-\obf_n^{(t)}\|_2^2 + \left(\frac{\lambda }{2Np}\sum_{n=1}^N\|\obf_n^{(t)}\|_2\right)^2}.
\end{equation}


\begin{algorithm}
\caption{Robust probabilistic clustering} \label{alg:RPC}
\begin{algorithmic}[1]
\REQUIRE Input data matrix $\mathbf{X}$, number of clusters $C$, and parameter $\lambda>0$.
\STATE Randomly initialize $\mathbf{M}^{(0)}$, $\boldsymbol{\pi}^{(0)}\in\mathcal{P}$, and set $\Sigmabf^{(0)}=\delta\Ibf_p$ ($\sigma^{(0)}=\sqrt{\delta}$) for $\delta>0$, and $\mathbf{O}^{(0)}$ to zero.
\FOR{$t=1,2,\ldots$}
\STATE Update $\gamma_{nc}^{(t)}$ via
\eqref{eq:gamma_update} for all $n,c$. 
\STATE Update $\boldsymbol{\pi}^{(t)}$ via \eqref{eq:pi_update}. 
\STATE Update $\mathbf{M}^{(t)}$ via \eqref{eq:m_updateEM}. 
\STATE Update $\mathbf{O}^{(t)}$ by solving \eqref{eq:o_minEM} (\eqref{eq:o_updateEM}).
\STATE Update $\Sigmabf^{(t)}$ ($\sigma^{(t)}$) via \eqref{eq:Sigma_update} (\eqref{eq:sigma_update}). 
\ENDFOR
\end{algorithmic}
\end{algorithm}

The robust probabilistic clustering (RPC) scheme is tabulated as Algorithm \ref{alg:RPC}. For spherical clusters, its complexity remains $\mathcal{O}(NCp)$ operations per iteration, even though the constants involved are larger than those in the RKM algorithm. Similar to RKM, the RPC iterations are convergent under mild conditions. Convergence of the RPC iterates is established in the next proposition. 

\begin{proposition}\label{pr:RPCconv}
The RPC iterations converge to a coordinate-wise minimum of the log-likelihood in \eqref{eq:likelihood}.
\end{proposition}

\begin{IEEEproof}
Combining the two steps of the EM algorithm, namely \eqref{eq:gamma_update} and \eqref{eq:EM_cost}, it is easy to verify that the algorithm is equivalent to a sequence of BCD iterations for optimizing 
\begin{equation}\label{eq:EMproof}
\min_{\mathbf{\Gamma},\mathbf{\Theta}'} ~ -~\sum_{n=1}^N\sum_{c=1}^C\gamma_{nc} \log\left(\frac{\pi_c \mathcal{N}\left(\xbf_n;\mbf_c+\obf_n,\Sigmabf\right)}{\gamma_{nc} }\right) + \lambda\sum_{n=1}^N\|\mathbf{o}_n\|_{\Sigmabf^{-1}} + f_s(\mathbf{\Gamma}\in \mathcal{U}_2)+ f_s(\boldsymbol{\pi}\in \mathcal{P}) + f_s(\mathbf{\Sigmabf}\succ0)
\end{equation}
where $\mathbf{\Theta}':=\{\boldsymbol{\pi},\Mbf,\Obf,\mathbf{\Sigma}\}$, the $N\times C$ matrix $\mathbf{\Gamma}$ has entries $[\mathbf{\Gamma}]_{n,c}:=\gamma_{nc}>0$, and as in \eqref{eq:equiv1} that $f_s(c)$ is zero when condition $c$ is true, and $\infty$ otherwise. That the $\{\gamma_{nc}\}$ are positive follows after using Bayes' rule to deduce that $\gamma_{nc}\propto \pi_c \mathcal{N}\left(\xbf_n;\mbf_c+\obf_n,\Sigmabf\right)$
and noticing that (i) $\mathcal{N}\left(\xbf_n;\mbf_c+\obf_n,\Sigmabf\right)$ is positive for all $\xbf_n$, and (ii) all $\pi_c$ must be positive so that the cost in \eqref{eq:EMproof} remains finite. 

The objective function of this minimization problem is proper, bounded below, and lower semi-continuous implying, that its non-empty level sets are closed. Since this function is also coercive, its level sets are bounded. Hence, its non-empty level sets are compact. Moreover, the objective function has a unique minimizer for the optimization blocks $\pibf$, $\Mbf$, and $\Obf$. In particular, the $\mathbf{M}$ block minimizer is unique since $\sum_{n=1}^N\gamma_{nc}>0$, for all $c\in\Natural_C$. Then, by \cite[Th.~4.1 (c)]{Ts01}, the RPC algorithm converges to a coordinate-wise minimum point of \eqref{eq:likelihood}.
\end{IEEEproof}

Proposition \ref{pr:RPCconv} guarantees that the RPC iterations converge. However, since each non-differentiable term $\|\obf_n\|_{\Sigmabf^{-1}}$ involves two different optimization variables $\Sigmabf$ and $\obf_n$, the BCD iteration can be trapped at a coordinate-wise local minimum, which is not necessarily a local minimum of \eqref{eq:robust_MLE}. Once the iterations have converged, the final $\gamma_{nc}$'s can be interpreted as soft cluster assignments, whereby hard assignments can be obtained via the maximum a posteriori detection rule, i.e., $\xbf_n\in \mathcal{X}_c$ for $c=\arg\max_{c'}\gamma_{nc'}$.

\begin{remark}[{\it Selecting $\lambda$}] \label{re:selectlambda}
Tuning $\lambda$ is possible if additional information, e.g., on the percentage of outliers, is available. The robust clustering algorithm is ran for a decreasing sequence of $\lambda$ values $\{\lambda_g\}$, using ``warm starts'' \cite{FrHa07}, until the expected number of outliers is identified. When solving for $\lambda_g$, warm start refers to the optimization variables initialized to the solution obtained for $\lambda_{g-1}$. Hence, running the algorithm over $\{\lambda_g\}$ becomes very efficient, because few BCD iterations per $\lambda_g$ suffice for convergence.
\end{remark}

\subsection{Weighted Robust Clustering Algorithms}\label{subsec:weighed}
As already mentioned, the robust clustering methods presented so far approximate the discontinuous penalty $\indicator(\|\obf_n\|_2 > 0)$ by $\|\obf_n\|_2$, mimicking the CS paradigm in which $\indicator(|x|> 0)$ is surrogated by the convex function $|x|$. However, it has been argued that non-convex functions such as $\log(|x|+\epsilon)$ for a small $\epsilon>0$ can offer tighter approximants of $\indicator(|x|> 0)$ \cite{WeEl03}. This rationale prompted us to replace $\|\obf_n\|_2$ in \eqref{eq:robust_hard}, \eqref{eq:robust_soft}, and \eqref{eq:robust_MLE}, by the penalty $\log(\|\obf_n\|_2+\epsilon)$ to further enhance block sparsity in $\mathbf{o}_n$'s, and thereby improve resilience to outliers.

Altering the regularization term modifies the BCD algorithms only when minimizing wrt $\mathbf{O}$. This particular step remains decoupled across $\mathbf{o}_n$'s, but instead of the $\phi^{(t)}(\mathbf{o}_n)$ defined in \eqref{eq:phi}, one minimizes 
\begin{equation}\label{eq:wphi}
\phi^{(t)}_w(\obf_n):=\sum_{c=1}^C (u_{nc}^{(t-1)})^q \left(\|\xbf_n-\mbf_c^{(t)}
-\obf_n\|_2^2+\lambda\log\left(\|\obf_n\|_2+\epsilon\right)\right)
\end{equation}
that is no longer convex. The optimization in \eqref{eq:wphi} is performed using a single iteration of the majorization-minimization (MM) approach\footnote{Note that the MM approach for minimizing $\phi^{(t)}_w(\obf_n)$ at the $t$-th BCD iteration involves several internal MM iterations. Due to the external BCD iterations and to speed up the algorithm, a single MM iteration is performed per BCD iteration $t$.} \cite{LaHuYa00}. The cost $\phi^{(t)}_w(\obf_n)$ is majorized by a function $f^{(t)}(\obf_n; \obf_n^{(t-1)})$, which means that $\phi^{(t)}_w(\obf_n) \leq f^{(t)} (\obf_n; \obf_n^{(t-1)})$ for every $\mathbf{o}_n$ and $\phi^{(t)}_w(\obf_n) = f^{(t)} (\obf_n; \obf_n^{(t-1)})$ when $\mathbf{o}_n=\obf_n^{(t-1)}$. Then $f^{(t)}(\obf_n; \obf_n^{(t-1)})$ is minimized wrt $\obf_n$ to obtain $\mathbf{o}_n^{(t)}$.

To find a majorizer for $\phi^{(t)}_w(\obf_n)$, the concavity of the logarithm is exploited, i.e., the fact that $\log x\leq \log x_o + x/x_o -1$ for any positive $x$ and $x_o$. Applying the last inequality for the penalty and ignoring the constant terms involved, we end up minimizing 
\begin{equation}\label{eq:mphi}
f^{(t)}(\obf_n; \obf_n^{(t-1)}):=\sum_{c=1}^C (u_{nc}^{(t-1)})^q \left(\|\xbf_n-\mbf_c^{(t)}
-\obf_n\|_2^2 + \lambda_n^{(t)} \|\obf_n\|_2\right)
\end{equation}
where $\lambda_n^{(t)}:=\lambda / (\|\mathbf{o}_n^{(t-1)}\|_2+\epsilon)$. Comparing \eqref{eq:mphi} to \eqref{eq:phi} shows that the new regularization results in a weighted version of the original one. The only difference between the robust algorithms presented earlier and their henceforth termed \emph{weighted} counterparts is the definition of $\lambda$. At iteration $t$, larger values for $\|\obf_n^{(t-1)}\|_2$ lead to smaller thresholds in the thresholding rules (cf. \eqref{eq:o_update}, \eqref{eq:o_updateEM}), thereby making $\mathbf{o}_n$ more likely to be selected as nonzero. The weighted robust clustering algorithms initialize $\mathbf{o}_n^{(0)}$ to the associated $\mathbf{o}_n$ value the non-weighted algorithm converged to. Thus, to run the weighted RKM for a specific value of $\lambda$, the RKM needs to be run first. Then, weighted RKM is run with all the variables initialized to the values attained by RKM, but with the $\lambda_n^{(1)}$ as defined earlier. 

The MM step combined with the BCD algorithms developed hitherto are convergent under mild assumptions. To see this, note that the sequences of objective values for the RKM and RPC algorithms are both non-increasing. Since the respective cost functions are bounded below, those sequences are guaranteed to converge. Characterizing the points and speed of convergence goes beyond the scope of this paper.

\section{Clustering High-Dimensional and Nonlinearly Separable Data}\label{sec:high_d}
The robust clustering algorithms of Section \ref{sec:solvers} are kernelized here. The advantage of kernelization is twofold: (i) yields computationally efficient algorithms when dealing with high-dimensional data, and (ii) robustly identifies nonlinearly separable clusters.

\subsection{Robust K-means for High-Dimensional Data}\label{subsec:RKM_hd}
The robust clustering algorithms derived so far involve generally $\mathcal{O}(NCp)$ operations per iteration. However, several applications entail clustering relatively few but \emph{high-dimensional} data in the presence of outliers. In imaging applications, one may wish to cluster $N=500$ images of say $p=800\times 600=480,000$ pixels; while in DNA microarray data analysis, some tens of (potentially erroneous or rarely occurring) DNA samples are to be clustered based on their expression levels over thousands of genes \cite{hastie09}. In such clustering scenarios where $p\gg N$, an efficient method should avoid storing and processing $p$-dimensional vectors \cite{Dhillon04kkmspecclust}. To this end, the algorithms of Section \ref{sec:solvers} are \emph{kernelized} here \cite{ScSm02}. It will be shown that these kernelized algorithms require $\mathcal{O}(N^3C)$ operations per iteration and $\mathcal{O}(N^2)$ space; hence, they are preferable when $p>N^2$. This kernelization not only offers processing savings in the high-dimensional data regime, but also serves as the building module for identifying nonlinearly separable data clusters as pursued in the next subsection.

We focus on kernelizing the robust soft K-means algorithm; the kernelized robust hard K-means can then be derived after simple modifications. Consider the $N\times C$ matrix $\mathbf{U}_q$ with entries $[\mathbf{U}_q]_{nc}:=u_{nc}^q$, and the Grammian $\mathbf{K}:=\mathbf{X}^T\mathbf{X}$ formed by all pairwise inner products between the input vectors. Even though the cost for computing $\mathbf{K}$ is $\mathcal{O}(N^2p)$, it is computed only once. Note that the updates \eqref{eq:m_update}, \eqref{eq:o_update}, and \eqref{eq:u_update} involve inner products between the $p$-dimensional vectors $\{\mathbf{o}_n, \mathbf{r}_n\}_{n=1}^N$, and $\{\mathbf{m}_c\}_{c=1}^C$. If $\{\mathbf{v}_i\in\mathbb{R}^p\}_{i=1}^2$ is a pair of any of these vectors, the cost for computing $\mathbf{v}_1^T\mathbf{v}_2$ is clearly $\mathcal{O}(p)$. But if at every BCD iteration the aforementioned vectors lie in $\range(\mathbf{X})$, i.e., if there exist $\{\mathbf{w}_i\in\mathbb{R}^N\}_{i=1}^2$ such that $\{\mathbf{v}_i=\mathbf{X}\mathbf{w}_i\}_{i=1}^2$, then $\mathbf{v}_1^T\mathbf{v}_2= \mathbf{w}_1^T\mathbf{K}\mathbf{w}_2$, and the inner product can be alternatively calculated in $\mathcal{O}(N^2)$.

Hinging on this observation, it is first shown that all the $p\times 1$ vectors involved indeed lie in $\range(\mathbf{X})$. The proof is by induction: if at the $(t-1)$-st iteration every $\mathbf{o}_n^{(t-1)}\in\range(\mathbf{X})$ and $\mathbf{U}^{(t-1)}\in\mathcal{U}_2$, it will be shown that $\mathbf{o}_n^{(t)}$, $\mathbf{m}_c^{(t)}$, $\mathbf{r}_n^{(t)}$ updated by RKM lie in $\range(\mathbf{X})$ as well.

Suppose that at the $t$-th iteration, the matrix $\mathbf{U}^{(t-1)}$ defining $\mathbf{U}_q^{(t-1)}$ is in $\mathcal{U}_2$, while there exists matrix $\mathbf{A}^{(t-1)}$ such that $\mathbf{O}^{(t-1)}=\mathbf{X}\mathbf{A}^{(t-1)}$. Then, the update of the centroids in \eqref{eq:m_update} can be expressed as
\begin{equation}\label{eq:m_update_hd}
\hspace{1.2cm}\mathbf{M}^{(t)}=( \mathbf{X} -\mathbf{O}^{(t-1)}) \mathbf{U}_q^{(t-1)}\diag^{-1}((\mathbf{U}_q^{(t-1)})^T\mathbf{1}_N)=\mathbf{X}\mathbf{B}^{(t)}
\end{equation}
where
\begin{equation}\label{eq:beta}
\mathbf{B}^{(t)}:=(\mathbf{I}_N  -\mathbf{A}^{(t-1)}) \mathbf{U}_q^{(t-1)}\diag^{-1}((\mathbf{U}_q^{(t-1)})^T\mathbf{1}_N).
\end{equation}
Before updating $\mathbf{O}^{(t)}$, the residual vectors $\{\mathbf{r}_n\}$ must be updated via \eqref{eq:r_update}. Concatenating the residuals in $\mathbf{R}^{(t)}:= [\mathbf{r}_1^{(t)}\ldots \mathbf{r}_N^{(t)}]$, the update in \eqref{eq:r_update} can be rewritten in matrix form as
\begin{equation}\label{eq:r_update_hd}
\hspace{1.6cm}\mathbf{R}^{(t)}=\mathbf{X} -\mathbf{M}^{(t)} (\mathbf{U}_q^{(t-1)})^T\diag^{-1}(\mathbf{U}_q^{(t-1)}\mathbf{1}_C)=\mathbf{X} \mathbf{\Delta}^{(t)}
\end{equation}
where 
\begin{equation}\label{eq:delta}
\mathbf{\Delta}^{(t)}:=\mathbf{I}_N -\mathbf{B}^{(t)}(\mathbf{U}_q^{(t-1)})^T \diag^{-1}(\mathbf{U}_q^{(t-1)}\mathbf{1}_C).
\end{equation}
From \eqref{eq:o_update}, every $\mathbf{o}_n^{(t)}$ is a scaled version of $\mathbf{r}_n^{(t)}$ and the scaling depends on $\|\mathbf{r}_n^{(t)}\|_2$. Based on \eqref{eq:r_update_hd}, the latter can be readily computed as $\|\mathbf{r}_n^{(t)}\|_2=\sqrt{(\boldsymbol{\delta}_n^{(t)})^T \mathbf{K} \boldsymbol{\delta}_n^{(t)}}=\|\boldsymbol{\delta}_n^{(t)}\|_{\mathbf{K}}$, where $\boldsymbol{\delta}_n^{(t)}$ stands for the $n$-th column of $\mathbf{\Delta}^{(t)}$. Upon applying the thresholding operator, one arrives at the update 
\begin{equation}\label{eq:o_update_hd}
\mathbf{O}^{(t)}=\mathbf{X}\mathbf{A}^{(t)}
\end{equation}
where the $n$-th column of $\mathbf{A}^{(t)}$ is given by
\begin{equation}\label{eq:alpha}
\boldsymbol{\alpha}_n^{(t)}=\boldsymbol{\delta}_n^{(t)}\left[1-\frac{\lambda}{2\|\boldsymbol{\delta}_n^{(t)}\|_{\mathbf{K}}}\right]_+,~~\forall n.
\end{equation}

Having proved the inductive step by \eqref{eq:o_update_hd}, the argument is complete if and only if the outlier variables $\mathbf{O}$ are initialized as $\mathbf{O}^{(0)}=\mathbf{X}\mathbf{A}^{(0)}$ for some $\mathbf{A}^{(0)}$, including the practically interesting and meaningful initialization at zero. The result just proved can be summarized as follows.

\begin{proposition}\label{pr:induction}
By choosing $\mathbf{O}^{(0)}=\mathbf{X}\mathbf{A}^{(0)}$ for any $\mathbf{A}^{(0)}\in\mathbb{R}^{N\times N}$ and $\mathbf{U}^{(0)}\in\mathcal{U}_2$, the columns of the matrix variables $\mathbf{O}$, $\mathbf{M}$, and $\mathbf{R}$ updated by RKM all lie in $\range(\mathbf{X})$; i.e., there exist known $\mathbf{A}^{(t)}$, $\mathbf{B}^{(t)}$, and $\mathbf{\Delta}^{(t)}$, such that $\mathbf{O}^{(t)}=\mathbf{X}\mathbf{A}^{(t)}$, $\mathbf{M}^{(t)}=\mathbf{X}\mathbf{B}^{(t)}$, and $\mathbf{R}^{(t)}=\mathbf{X}\mathbf{\Delta}^{(t)}$ for all $t$.
\end{proposition}

What remains to be kernelized are the updates for the cluster assignments. For the update step \eqref{eq:u_update} or \eqref{eq:u_update_hard}, we need to compute $\|\xbf_n-\mbf_c^{(t)}-\obf_n^{(t)}\|_2^2$ and $\|\obf_n^{(t)}\|_2$. Given that $\mathbf{x}_n=\mathbf{X}\mathbf{e}_n$, where $\mathbf{e}_n$ denotes the $n$-th column of $\mathbf{I}_N$, and based on the kernelized updates \eqref{eq:m_update_hd} and \eqref{eq:o_update_hd}, it is easy to verify that
\begin{equation}\label{eq:norm2_2_hd}
\|\xbf_n-\mbf_c^{(t)}-\obf_n^{(t)}\|_2^2= \|\mathbf{X}(\mathbf{e}_n-\boldsymbol{\beta}_{c}^{(t)}-\boldsymbol{\alpha}_n^{(t)})\|_2^2 = \|\mathbf{e}_n-\boldsymbol{\beta}_{c}^{(t)}-\boldsymbol{\alpha}_n^{(t)}\|_{\mathbf{K}}^2
\end{equation}
for every $n$ and $c$, where $\boldsymbol{\beta}_c^{(t)}$ is the $c$-th column of $\mathbf{B}^{(t)}$. As in \eqref{eq:o_update_hd}, it follows that
\begin{equation}\label{eq:norm2_hd}
\|\obf_n^{(t)}\|_2= \|\mathbf{X}\boldsymbol{\alpha}_n^{(t)}\|_2= \|\boldsymbol{\alpha}_n^{(t)}\|_{\mathbf{K}}.
\end{equation}

The kernelized robust K-means (KRKM) algorithm is summarized as Algorithm \ref{alg:KRKM}. As for RKM, the KRKM algorithm is terminated when $\|\Mbf^{(t)}-\Mbf^{(t-1)}\|_{F}/\|\Mbf^{(t)}\|_{F} \leq \epsilon_s$ for a small $\epsilon_s>0$. Based on \eqref{eq:m_update_hd} and exploiting standard linear algebra properties, the stopping condition can be equivalently expressed as $(\sum_{c=1}^C \|\boldsymbol{\beta}_c^{(t)} - \boldsymbol{\beta}_c^{(t-1)}\|_{\mathbf{K}}^2)/
(\sum_{c=1}^C \|\boldsymbol{\beta}_c^{(t)}\|_{\mathbf{K}}^2)\leq\epsilon_s^2$.

Notice that this kernelized algorithm does not explicitly update $\mathbf{M}$, $\mathbf{R}$, or $\mathbf{O}$; actually, these variables are never processed. Instead, it updates $\mathbf{A}$, $\mathbf{B}$, and $\mathbf{\Delta}$; while the clustering assignments are updated via \eqref{eq:u_update}, \eqref{eq:norm2_2_hd}, and \eqref{eq:norm2_hd}. Ignoring the cost for finding $\mathbf{K}$, the computations required by this algorithm are $\mathcal{O}(N^3C)$ per iteration, whereas the stored variables $\mathbf{A}$, $\mathbf{B}$, $\mathbf{\Delta}$, $\mathbf{U}_q$, and $\mathbf{K}$ occupy $\mathcal{O}(N^2)$ space. Note that if the centroids $\mathbf{M}$ are explicitly needed  (e.g., for interpretative purposes or for clustering new input data ), they can be acquired via \eqref{eq:m_update_hd} after KRKM has terminated.

\begin{algorithm}
\caption{Kernelized RKM }\label{alg:KRKM}
\begin{algorithmic}[1]
\REQUIRE Grammian matrix $\mathbf{K}\succ 0$, number of clusters $C$, $q\geq1$, and $\lambda>0$.
\STATE Initialize $\Ubf^{(0)}$ randomly in $\mathcal{U}_2$, and $\mathbf{A}^{(0)}$ to zero.  
\FOR{$t=1,2,\ldots$}
\STATE Update $\mathbf{B}^{(t)}$ from \eqref{eq:beta}.
\STATE Update $\mathbf{\Delta}^{(t)}$ from \eqref{eq:delta}.
\STATE Update $\mathbf{A}^{(t)}$ from \eqref{eq:alpha}.
\STATE Update $\mathbf{U}^{(t)}$ and $\mathbf{U}_q^{(t)}$ from \eqref{eq:u_update} or \eqref{eq:u_update_hard}, \eqref{eq:norm2_2_hd}, and \eqref{eq:norm2_hd}.
\ENDFOR
\end{algorithmic}
\end{algorithm}

\subsection{Kernelized RKM for Nonlinearly Separable Clusters}\label{subsec:RKM_kernel}
One of the limitations of conventional K-means is that clusters should be of spherical or, more generally, ellipsoidal shape. By postulating the squared Euclidean distance as the similarity metric between vectors, the underlying clusters are tacitly assumed to be linearly separable. GMM-based clustering shares the same limitation. Kernel K-means has been proposed to overcome this obstacle \cite{ScSmMu98} by mapping vectors $\xbf_n$ to a higher dimensional space $\mathcal{H}$ through the nonlinear function $\varphi:\mathbb{R}^p\rightarrow \mathcal{H}$. The mapped data $\{\varphi(\mathbf{x}_n)\}_{n=1}^N$ lie in the so-termed feature space which is of dimension $P>p$ or even infinite. The conventional K-means algorithm is subsequently applied in its kernelized version on the transformed data. Thus, linearly separable partitions in feature space yield nonlinearly separable partitions in the original space.

For an algorithm to be kernelizable, that is to be able to operate with inputs in feature space, the inner product between any two mapped vectors, i.e., $\varphi^T(\mathbf{x}_n)\varphi(\mathbf{x}_m)$, should be known. For the non-kernelized versions of K-means, RKM, and RPC algorithms, where the linear mapping $\varphi(\mathbf{x}_n)=\mathbf{x}_n$ can be trivially assumed, these inner products are directly computable and stored at the $(n,m)$-th entry of the Grammian matrix $\mathbf{K}=\mathbf{X}^T\mathbf{X}$. When a nonlinear mapping is used, the so-termed kernel matrix $\mathbf{K}$ with entries $[\mathbf{K}]_{n,m}:=\varphi^T(\mathbf{x}_n)\varphi(\mathbf{x}_m)$ replaces the Grammian matrix and must be known. By definition, $\mathbf{K}$ must be positive semidefinite and can be employed for (robust) clustering, even when $\varphi(\mathbf{x}_n)$ is high-dimensional (cf. Section \ref{subsec:RKM_hd}), infinite-dimensional, or even unknown \cite{DhGuKu07}. 

Of particular interest is the case where $\mathcal{H}$ is a reproducing kernel Hilbert space. Then, the inner product in $\mathcal{H}$ is provided by a known kernel function $\kappa(\mathbf{x}_n,\mathbf{x}_m):= \varphi^T(\mathbf{x}_n)\varphi(\mathbf{x}_m)$ \cite[Ch. 3]{ScSm02}. Typical kernels for vectorial input data are the polynomial, Gaussian, and the sigmoid ones; however, kernels can be defined for non-vectorial objects as well, such as strings or graphs \cite{ScSm02}.

After having RKM tailored to the high-dimensional input data regime in Section \ref{subsec:RKM_hd}, handling arbitrary kernel functions is now straightforward. Knowing the input data $\mathbf{X}$ and the kernel function $\kappa(\mathbf{x}_n,\mathbf{x}_m)$, the kernel matrix can be readily computed as $[\mathbf{K}]_{n,m}=\kappa(\mathbf{x}_n,\mathbf{x}_m)$ for $n,~m\in \Natural_N$. By using the kernel in lieu of the Grammian matrix, Algorithm \ref{alg:KRKM} carries over readily to the nonlinear clustering regime.

As shown earlier, when clustering high-dimensional data, the centroids can be computed after the robust clustering algorithm has terminated via \eqref{eq:m_update_hd}. This is not generally the case withrobust nonlinear clustering. For infinite-dimensional or even finite dimensional feature spaces, such as the one induced by a polynomial kernel, even if one is able to recover the centroid $\mathbf{m}_c\in\mathcal{H}$, its pre-image in the input space may not exist \cite[Ch.~18]{ScSm02}.

\subsection{Kernelized Robust Probabilistic Clustering}\label{subsec:RPC_hd}
Similar to RKM, the RPC algorithm can be kernelized to (i) facilitate computationally efficient clustering of high-dimensional input data, and (ii) enable nonlinearly separable clustering. To simplify the presentation, the focus here is on the case of spherical clusters.

Kernelizing RPC hinders a major difference over the kernelization of RKM: the GMM and the RPC updates in Section \ref{subsec:EMsolver} remain valid for $\{\varphi(\xbf_n)\in \mathcal{H}\}$ only when the feature space dimension $P$ remains finite and known. The implication is elucidated as follows. First, updating the variance in \eqref{eq:sigma_update} entails the underlying vector dimension $p$ -- which becomes $P$ when it comes to kernelization. Second, the (outlier-aware) mixtures of Gaussians degenerate when it comes to modeling infinite-dimensional random vectors. To overcome this limitation, the notion of the empirical kernel map will be exploited \cite[Ch. 2.2.6]{ScSm02}. Given the fixed set of vectors in $\Xcal$, instead of $\varphi$, it is possible to consider the empirical kernel map $\hat{\varphi}:\Real^p\rightarrow\Real^N$ defined as $\hat{\varphi}(\xbf):=(\mathbf{K}^{1/2})^{\dagger}[\kappa(\xbf_1,\xbf)\cdots\kappa(\xbf_N,\xbf)]^T$, where $\mathbf{K}$ is the kernel matrix of the input data $\Xcal$, and $(\cdot)^\dagger$ the Moore-Penrose pseudoinverse. The feature space $\hat\Hcal$ induced by $\hat\varphi$ has finite dimensionality $N$. It can be also verified that $\hat{\varphi}^T(\xbf_n)\hat{\varphi}(\xbf_m) = {\varphi}^T(\xbf_n){\varphi}(\xbf_m)=\kappa(\mathbf{x}_n,\mathbf{x}_m)$ for all $\xbf_n,~\xbf_m\in\Xcal$; hence, inner products in $\hat\Hcal$ are readily computable through $\kappa$.

In the kernelized probabilistic setup, vectors $\{\hat{\varphi}(\xbf_n)\}_{n=1}^N$ are assumed drawn from a mixture of $C$ multivariate Gaussian distributions with common covariance $\Sigmabf=\sigma^2\Ibf_N$ for all clusters. The EM-based updates of RPC in Section \ref{subsec:EMsolver} remain valid after replacing the dimension $p$ in \eqref{eq:sigma_update} by $N$, and the input vectors $\{\xbf_n\}$ by $\{\hat{\varphi}(\xbf_n)\}$ with the critical property that one only needs to know the inner products $\hat{\varphi}^T(\mathbf{x}_n)\hat{\varphi}(\mathbf{x}_m)$ stored as the $(n,m)$-th entries of the kernel matrix $\mathbf{K}$. The kernelization procedure is similar to the one followed for RKM: first, the auxiliary matrices $\mathbf{A}^{(t)}$, $\mathbf{B}^{(t)}$, and $\mathbf{\Delta}^{(t)}$ are introduced. By randomly initializing $\sigma^{(0)}$, $\boldsymbol{\pi}^{(0)}\in\mathcal{P}$, $\mathbf{B}^{(0)}\in\mathbb{R}^{N\times C}$, and setting $\mathbf{A}^{(0)}$ to zero, it can be shown as in Proposition \ref{pr:induction}, that the kernelized RPC updates for $\mathbf{O}^{(t)}$, $\mathbf{M}^{(t)}$, and $\mathbf{R}^{(t)}$ have their columns lying in $\range(\mathbf{\Phi})$, where $\mathbf{\Phi}:=[\hat{\varphi}(\mathbf{x}_1)\cdots\hat{\varphi}(\mathbf{x}_N)]$. Instead of the assignment matrix $\mathbf{U}$ in KRKM, the $N\times C$ matrix of posterior probability estimates $\mathbf{\Gamma}^{(t)}$ is used, where $[\mathbf{\Gamma}^{(t)}]_{n,c}:=\gamma_{nc}^{(t)}$ satisfying $\mathbf{\Gamma}^{(t)}\mathbf{1}_C=\mathbf{1}_N$ $\forall t$.


\begin{algorithm}
\caption{Kernelized RPC } \label{alg:KRPC}
\begin{algorithmic}[1]
\REQUIRE Grammian or kernel matrix $\mathbf{K}\succ 0$, number of clusters $C$, and $\lambda>0$.
\STATE Randomly initialize $\sigma^{(0)}$, $\boldsymbol{\pi}^{(0)}\in\mathcal{P}$, and $\mathbf{B}^{(0)}$; and set $\mathbf{A}^{(0)}$ to zero.
\FOR{$t=1,2,\ldots$}
\STATE Update $\mathbf{\Gamma}^{(t)}$ via \eqref{eq:gamma_update} exploiting $\|\xbf_n-\mbf_c^{(t-1)}-\obf_n^{(t-1)}\|_2^2=\|\mathbf{e}_n-\boldsymbol{\beta}_{c}^{(t-1)}-\boldsymbol{\alpha}_n^{(t-1)}\|_{\mathbf{K}}^2$ for all $n,c$.
\STATE Update $\boldsymbol{\pi}^{(t)}$ as $\boldsymbol{\pi}^{(t)}=(\mathbf{\Gamma}^{(t)})^T\mathbf{1}_N/N$.
\STATE Update $\mathbf{B}^{(t)}$ as $\mathbf{B}^{(t)}=(\mathbf{I}_N  -\mathbf{A}^{(t-1)}) \mathbf{\Gamma}^{(t)}\diag^{-1}(N\boldsymbol{\pi}^{(t)})$.
\STATE Update $\mathbf{\Delta}^{(t)}$ as $\mathbf{\Delta}^{(t)}=\mathbf{I}_N  -\mathbf{B}^{(t)} (\mathbf{\Gamma}^{(t)})^T$.
\STATE Update the columns of $\mathbf{A}^{(t)}$ as $\boldsymbol{\alpha}_n^{(t)}=\boldsymbol{\delta}_n^{(t)}\left[1-\frac{\lambda\sigma^{(t-1)}}{\|\boldsymbol{\delta}_n^{(t)}\|_{\mathbf{K}}}\right]_+$ for all $n$.
\STATE Update $\sigma^{(t)}$ via \eqref{eq:sigma_update} where $p$ is replaced by $N$, using the $\ell_2$-norms computed in Step 3, and exploiting $\|\mathbf{o}_n^{(t)}\|_2 = \|\boldsymbol{\alpha}_n^{(t)}\|_{\mathbf{K}}$ for all $n$.
\ENDFOR
\end{algorithmic}
\end{algorithm}

The kernelized RPC (KRPC) algorithm is summarized as Alg.~\ref{alg:KRPC}. As with KRKM, its computations are $\mathcal{O}(N^3C)$ per iteration, whereas the stored variables $\mathbf{A}$, $\mathbf{B}$, $\mathbf{\Delta}$, $\mathbf{\Gamma}$, $\boldsymbol{\pi}$, $\mathbf{K}$, and $\sigma$ occupy $\mathcal{O}(N^2)$ space.

\begin{remark}[{\it Reweighted kernelized algorithms}]
Reweighted kernelized algorithms similar to the ones in Section \ref{subsec:weighed} can be derived for KRKM and KRPC by a simple modification. In both cases, it suffices to introduce an iteration-dependent parameter $\lambda_n^{(t)}=\lambda/(\|\obf_n^{(t-1)}\|_2+\epsilon)$ per index $n$. Note that $\|\obf_n\|_2$'s can be readily computed in terms of kernels as shown earlier. 
\end{remark}

\section{Numerical Tests}\label{sec:tests}
Numerical tests illustrating the performance of the novel robust clustering algorithms on both synthetic and real datasets are shown in this section. Performance is assessed through their ability to identify outliers and the quality of clustering itself. The latter is measured using the adjusted rand index (ARI) between the partitioning found and the true partitioning of the data whenever the latter is available \cite{hubert85ARI}. In each experiment, the parameter $\lambda$ is tuned using the grid search outlined in Remark \ref{re:selectlambda} over at most $1,000$ values. Thanks to the warm-start technique, the solution path for all grid points was computed in an amount of time comparable to the one used for solving for a specific value of $\lambda$.

\subsection{Synthetic Datasets}
\begin{figure}
\centering
\subfigure[Dataset with $C=4$ spherical clusters.]{
\includegraphics[width=0.45\linewidth]{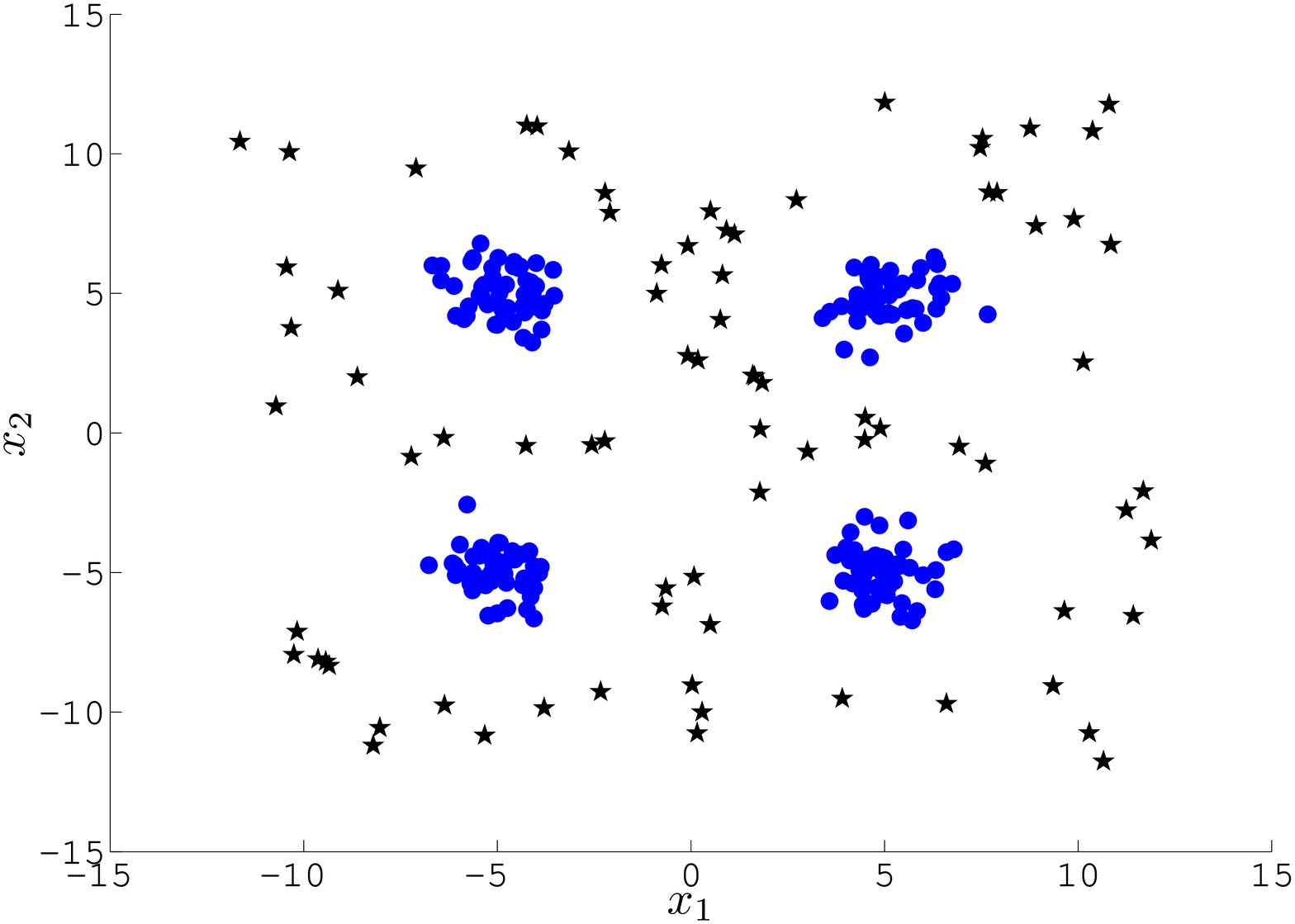}
\label{subfig:spherical}}
\subfigure[Dataset with $C=2$ concentric rings.]{
\includegraphics[width=0.45\linewidth]{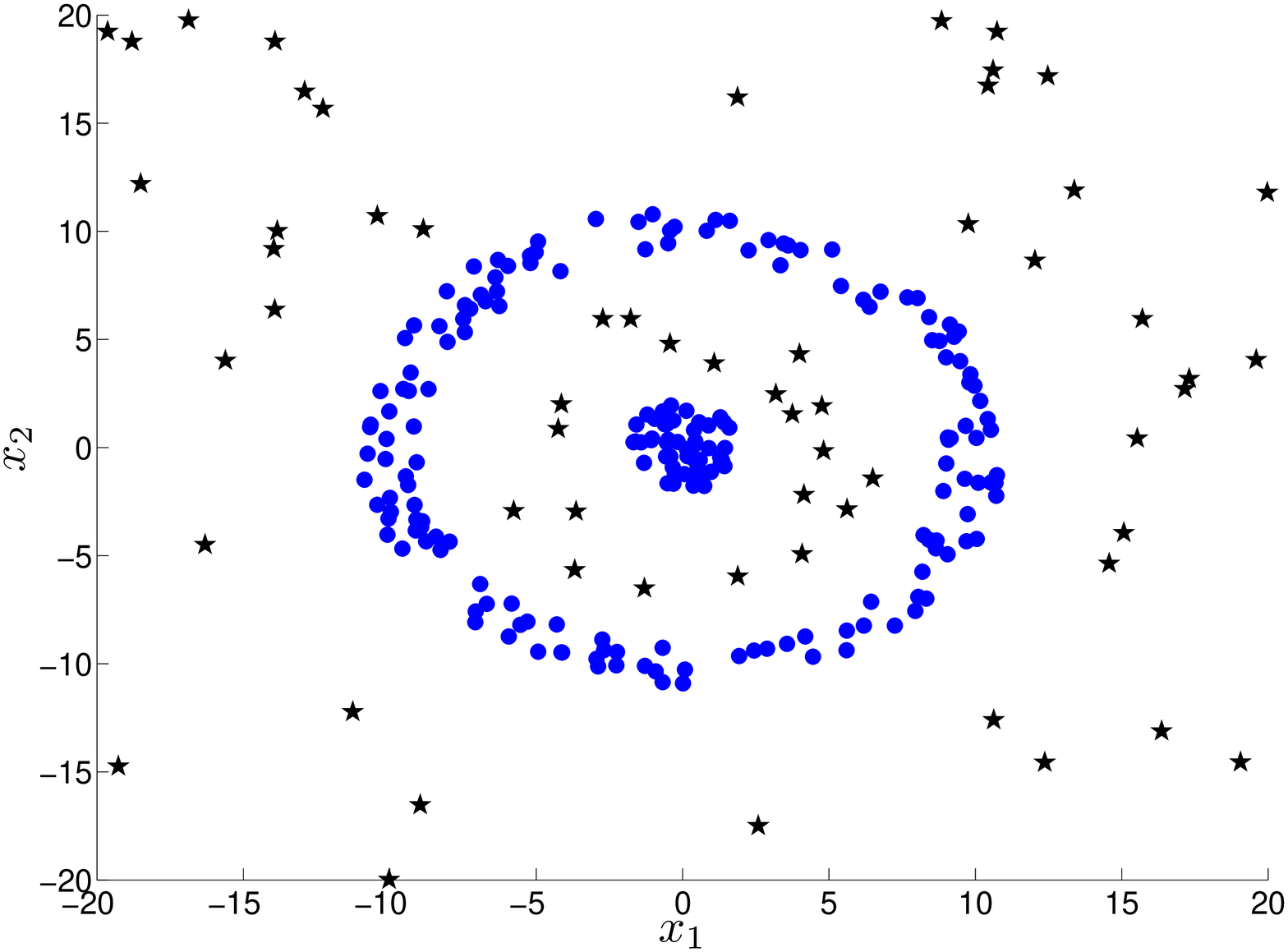}
\label{subfig:rings}}
\caption{Synthetic datasets: In-(out-)lier vectors are denoted by circles $\bullet$ (stars $\bigstar$).}
\label{fig:synthetic}
\end{figure}

Two synthetic datasets are used. The first one, shown in Fig.~\ref{subfig:spherical}, consists of a random draw of 200 vectors from $C=4$ bivariate Gaussian distributions (50 vectors per distribution), and 80 outlying vectors $(N=280)$. The Gaussian distributions have different means and a common covariance matrix $0.8\Ibf_2$. The second  dataset comprises points belonging to $C=2$ concentric rings as depicted in Fig.~\ref{subfig:rings}. The inner (outer) ring has 50 (150) points. It also contains $60$ vectors lying in the areas between the rings and outside the outer ring corresponding to outliers $(N=260)$. Clustering this second dataset is challenging even if outliers were not present due to the shape and multiscale nature of the clusters.

\begin{figure}
\centering
\subfigure[RKM algorithm for $q=1$.]{
\includegraphics[width=0.45\linewidth]{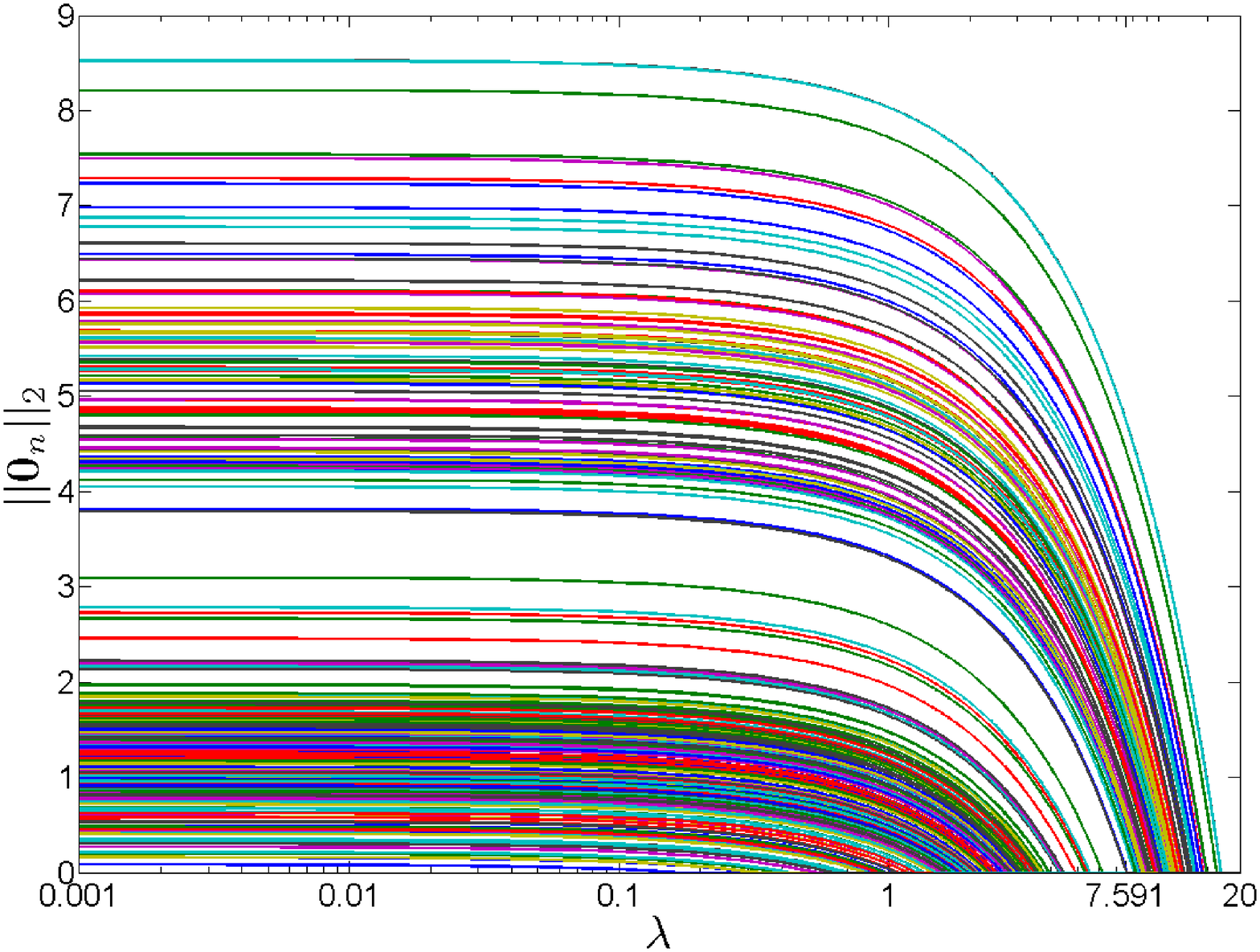}
\label{subfig:cluster4curves_RKM}}
\subfigure[RPC algorithm.]{
\includegraphics[width=0.45\linewidth]{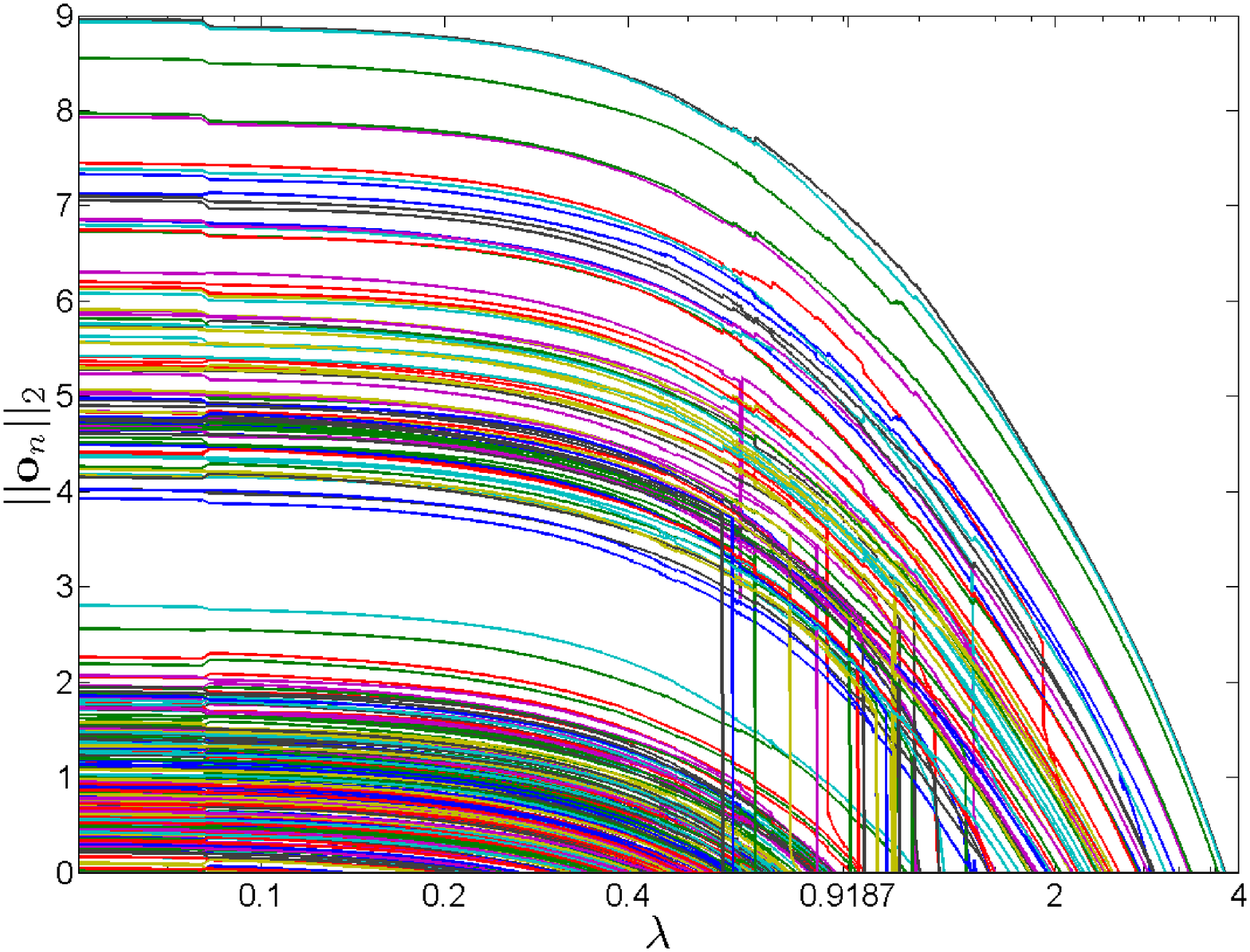}
\label{subfig:cluster4curves_RPC}}
\caption{Curves of $\|\obf_n\|_2$'s as a function of $\lambda$ for the dataset in Fig. \ref{subfig:spherical}.}
\label{fig:cluster4curves}
\end{figure}

Starting from the dataset with the four spherical clusters, the effect of $\lambda$ on the number of outliers identified is investigated. In Fig.~\ref{fig:cluster4curves}, the values of $\{\|\obf_n\|_2\}_{n=1}^N$ are plotted as a function of $\lambda$. The outlier-norm curves shown in Fig.~\ref{subfig:cluster4curves_RKM} correspond to the RKM algorithm with $q=1$ using a random initialization. For $\lambda>17$, all $\obf_n$'s are set to zero, while as $\lambda$ approaches zero, more $\obf_n$'s take nonzero values. Selecting $\lambda\in[6.2,7.6]$ yields 80 outliers. Fig.~\ref{subfig:cluster4curves_RPC} shows $\{\left\|\obf_n\right\|_2\}_{n=1}^N$ as $\lambda$ varies for the RPC algorithm assuming $\Sigmabf=\sigma^2\Ibf_p$. The curves for some $\obf_n$'s exhibit a fast transition from zero.


\begin{figure}
\centering
\includegraphics[width=0.5\linewidth]{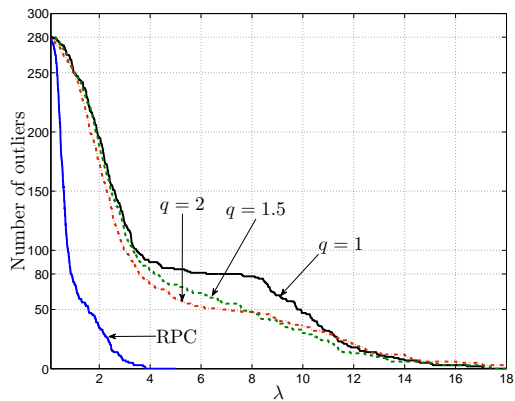}
\caption{Number of outliers identified as a function of $\lambda$ for the dataset in Fig. \ref{subfig:spherical}.}
\label{fig:NoO_spherical}
\end{figure}

In Fig.~\ref{fig:NoO_spherical}, the number of outliers identified, i.e., the number of input vectors $\mathbf{x}_n$ with corresponding $\|\obf_n\|_2>0$, is plotted as a function of $\lambda$. Proper choice of $\lambda$ enables both RKM and RPC to identify exactly the 80 points, which were generated as outliers in the ``ground truth'' model. For the RKM algorithm with $q=1$, there is a plateau for values of $\lambda\in[6.2,7.6]$. This plateau defines a transition region between the values of $\lambda$ that identify true outliers and values of $\lambda$ which erroneously deem non-outliers as outliers. Although the plateau is not present for $q>1$, the curves show an increase in their slope for $\lambda<5$ indicating that non-outliers are erroneously labeled as outliers. RPC with $\lambda=0.91$ correctly identifies the 80 outlying vectors. Notice that the range of $\lambda$ values for which outliers are correctly identified is smaller than the one for RKM due to the scaling of $\lambda$ by $\sigma$.

Table~\ref{table:t0} shows the root-mean-squared error (RMSE) of cluster center estimates averaged over 100 algorithm initializations. Two levels of outlier contamination are considered: 40 out of $N=240$ points (approximately 17\%), and 80 out of $N=280$ (approximately 29\%). Apart from the novel algorithms, hard K-means, soft K-means with $q=1.5$, and EM are also included. The robust versions of the hard K-means, soft K-means, and EM algorithms achieve lower RMSE with extra improvement effected by the weighted RKM (WRKM) and the weighted RPC (WRPC) algorithms.

\begin{table}[!t]
\renewcommand{\arraystretch}{1.3}
\caption{RMSE of cluster center estimates.}
\vspace*{-1em}
\label{table:t0}
\centering
\begin{tabular}{c|cc}
\hline
&\multicolumn{2}{c}{\bf RMSE}\\
\hline
\bf Outliers/$N$&{\bf40}/240&{\bf80}/280\\
\hline
hard K-means&0.7227&1.0892\\
soft K-means&0.5530&1.0206\\
EM&0.6143&1.0032\\
\hline
hard RKM&0.4986&0.8813\\
hard WRKM&0.4985&0.8812\\
soft RKM&0.2587&0.6259\\
soft WRKM&0.0937&0.1758\\
RPC&0.2789&0.3891\\
WRPC&0.1525&0.1750\\
\hline
\end{tabular}
\end{table}

\begin{figure}
\centering
\subfigure[KRKM algorithm.]{
\includegraphics[width=0.45\linewidth]{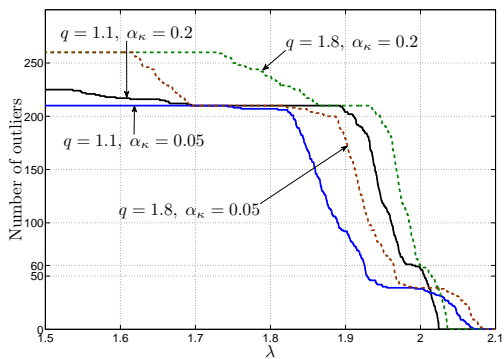}
\label{subfig:donut2curves_RKM}}
\subfigure[KRPC algorithm.]{
\includegraphics[width=0.45\linewidth]{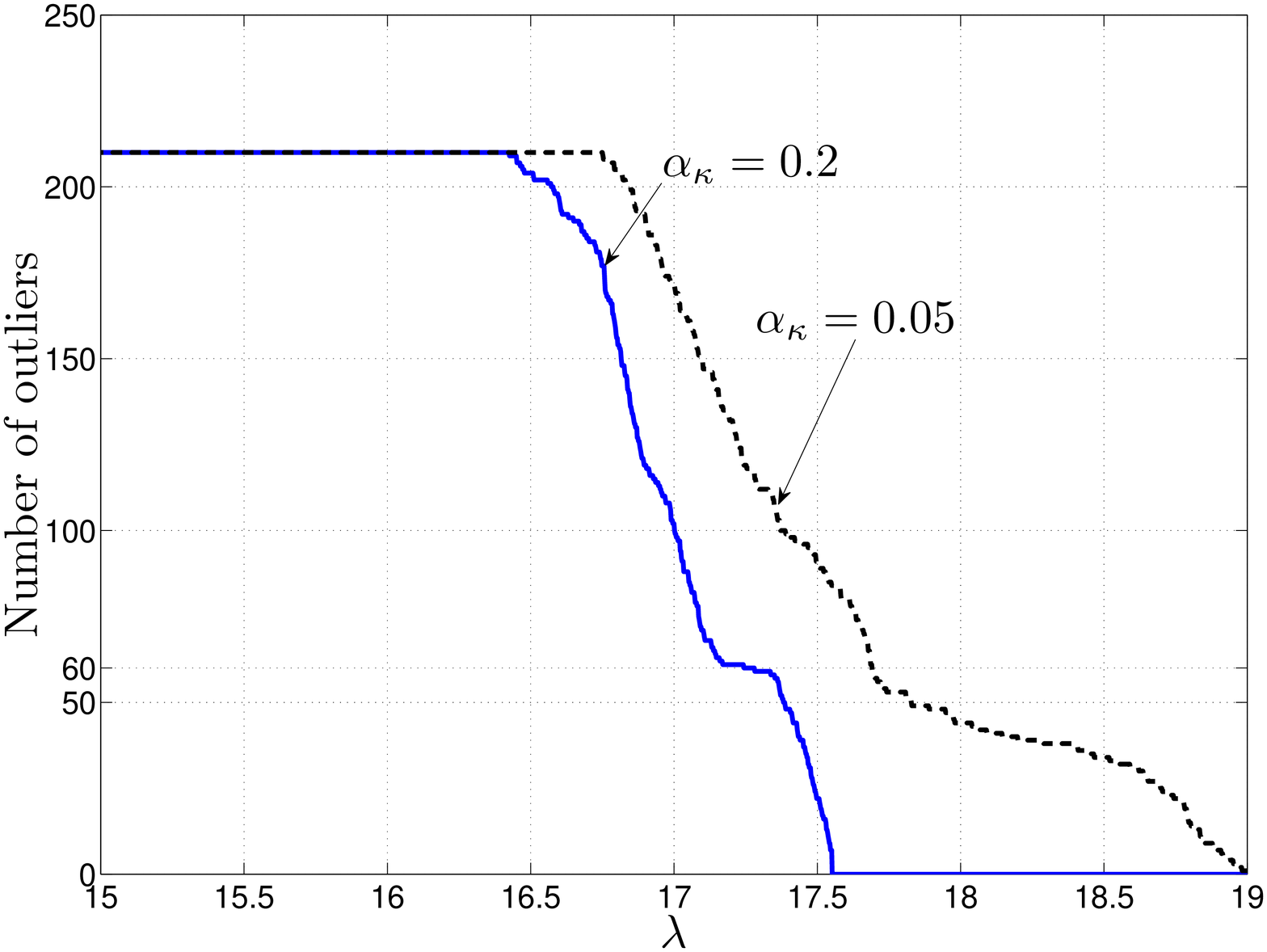}
\label{subfig:donut2curves_RPC}}
\caption{Number of outliers identified as a function of $\lambda$ for the dataset in Fig. \ref{subfig:rings}.}
\label{fig:donut2curves}
\end{figure}

\begin{figure}
\centering
\subfigure[KRKM algorithm $(q=1.1)$.]{
\includegraphics[width=0.45\linewidth]{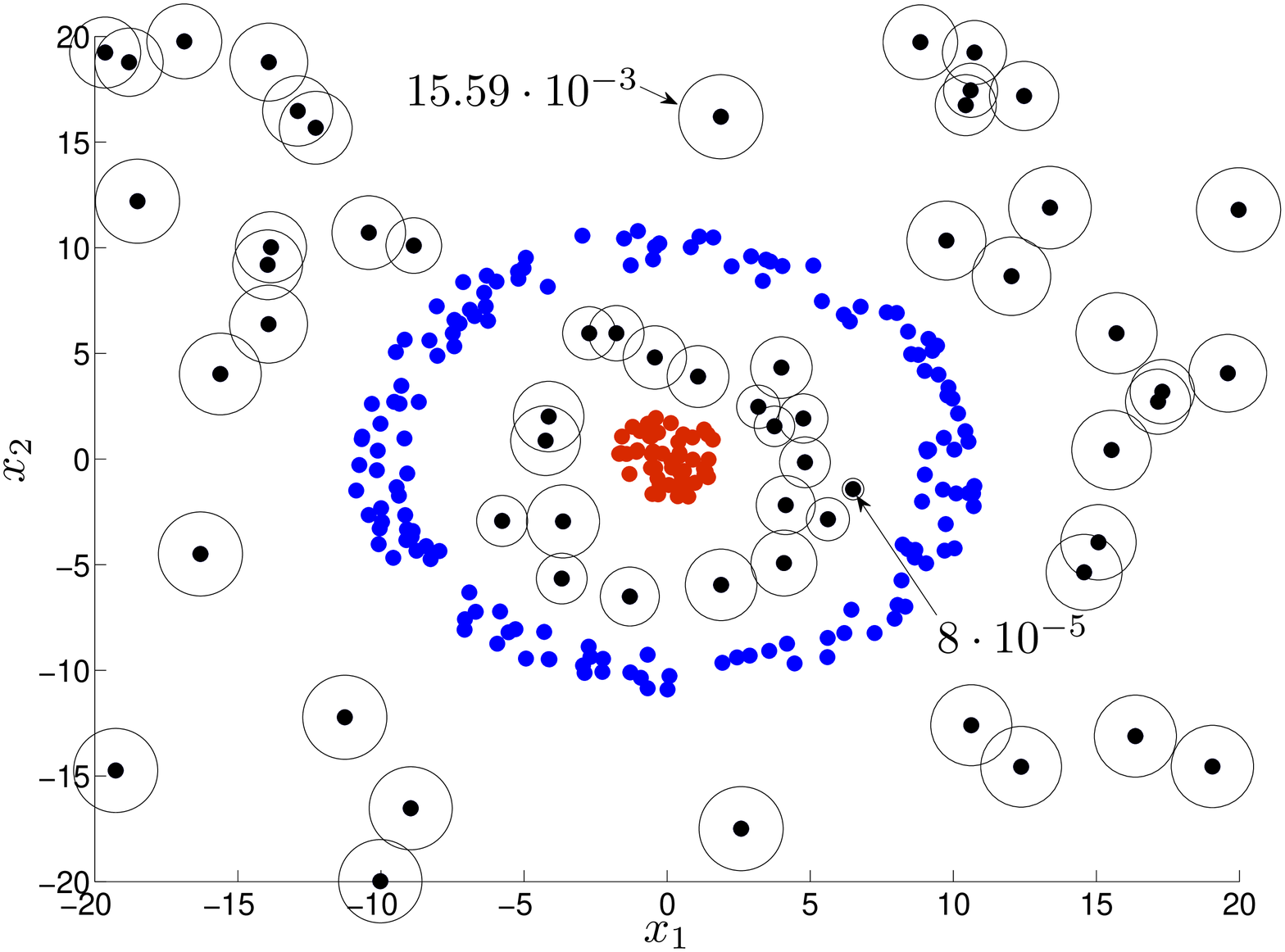}
\label{subfig:rings_outliers}}
\subfigure[KRPC algorithm.]{
\includegraphics[width=0.45\linewidth]{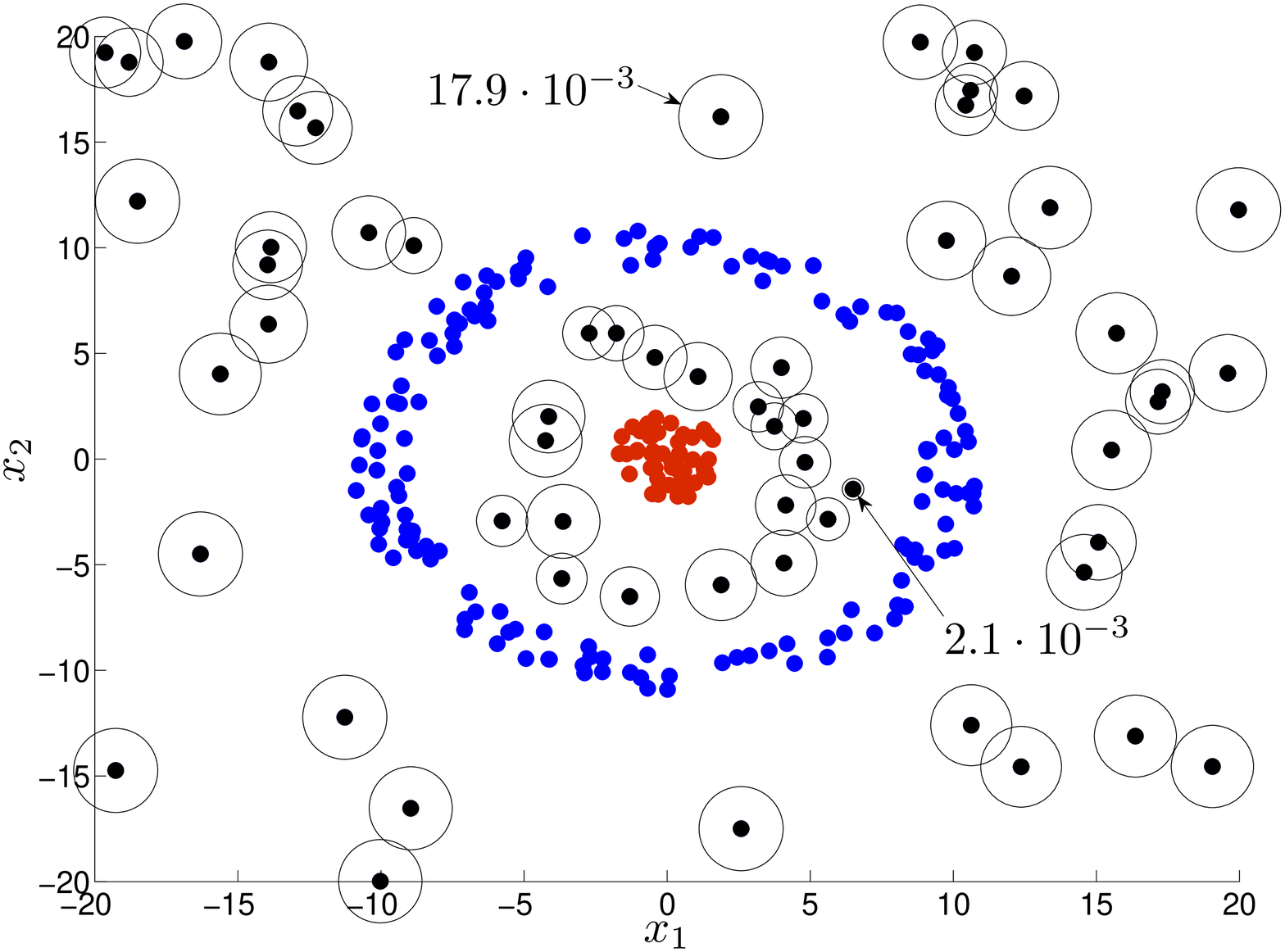}
\label{subfig:rings_outliersEM}}
\caption{Clustering results for the dataset in Fig. \ref{subfig:rings} using a Gaussian kernel with $\alpha_{\kappa}=0.2$. Points surrounded by a circle were deemed as outliers; the radius of the circle is proportional to the value of $\|\obf_n\|_2$. Smallest and largest $\|\obf_n\|_2$ values are shown.} 
\label{fig:rings_outliers}
\end{figure}

Next, we consider clustering the second nonlinearly separable dataset using the Gaussian kernel $\kappa(\xbf_n,\xbf_m)=\exp(-\alpha_{\kappa}\|\xbf_n-\xbf_m\|_2^2)$, where $\alpha_{\kappa}>0$ is a scaling parameter. The parameter $\alpha_{\kappa}^{-1}$ is chosen as a robust variance estimate of the entire dataset as described in \cite{chen09outliers}. Both KRKM and KRPC are able to identify the 60 outlying points. In Fig.~\ref{fig:donut2curves}, the number of outliers identified by KRKM and KRPC is plotted as a function of $\lambda$ for different values of $\alpha_{\kappa}$. Fig.~\ref{fig:rings_outliers} illustrates the values of $\|\obf_n\|_2$'s for WKRKM and WKRPC when seeking 60 outliers. Points surrounded by a circle correspond to vectors identified as outliers, and each circle's radius is proportional to its corresponding $\|\obf_n\|_2$ value.

\subsection{USPS Dataset}
In this subsection, the robust clustering algorithms are tested on the United States Postal Service (USPS) handwritten digit recognition corpus. This corpus contains gray-scale digit images of $16\times 16$ pixels with intensities normalized to $[-1,1]$. It is divided to 7,201 training and 2,007 test examples of the digits 0-9. Although the corpus contains class labels, they are known to be inconsistent: some digits are erroneously labeled, while some images are difficult to be classified even by humans \cite[App.~A]{ScSm02}. In this experiment, the subset of digits 0-5 is used. For each digit, both training and test sets are combined to a single set and then 300 images are sampled uniformly at random, yielding a dataset of 1800 images. Each image is represented as a 256-dimensional vector normalized to have unit $\ell_2$-norm.

Hard RKM $(q=1)$ and RPC algorithms are used to partition the dataset into $C=6$ clusters and identify $s=100$ outliers. All algorithms were tested for 20 Monte Carlo runs with random initializations common to all algorithms. The final partitioning is chosen as the one attaining the smallest cost in \eqref{eq:robust_hard}. The quality of the clustering is assessed through the ARI after excluding the outlier vectors. The ARI values for K-means, K-medians, and the proposed schemes are shown in Table~\ref{table:t1}. Note that the ARI values for RKM (RPC) and WRKM (WRPC) are equal. This indicates that the weighted algorithms do not modify the point-to-cluster assignments already found. Interestingly, the K-medians algorithm was not able to find a partitioning of the data revealing the 6 digits present, even after 100 Monte Carlo runs.

\begin{table}
\caption{ARI coefficients for the USPS dataset $(C=6)$} \centering
\begin{tabular}{c  |c c c c c}
\hline
Kernel & {\bf K-means} & {\bf K-medians} & {\bf RKM} & {\bf RPC} \\
\hline %
{Linear} &0.6469& 0.5382& 0.6573 & 0.6508\\
{Polynomial} &0.5571&-&0.6978&0.6965\\
\hline
\end{tabular}\label{table:t1}
\end{table}

\begin{figure}
\centering
\subfigure[RPC and WRPC centroids.]{
\includegraphics[scale=0.35]{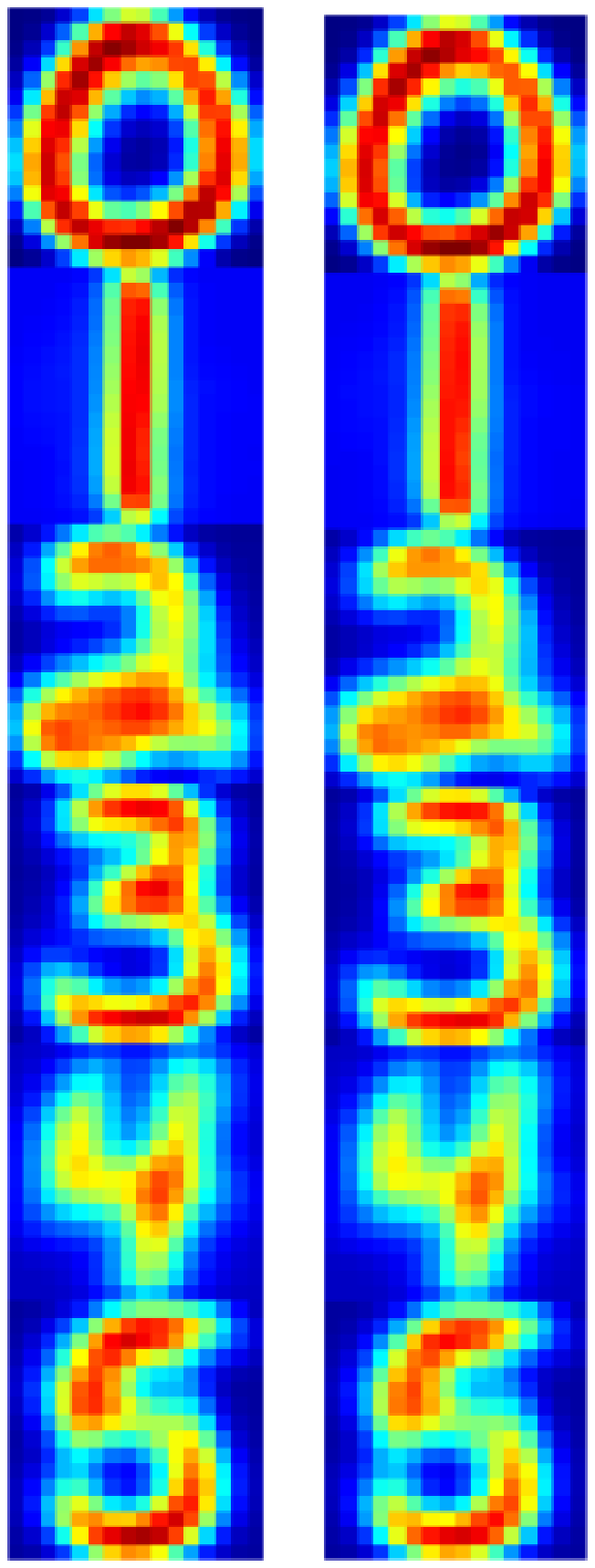}\label{fig:sim8_left}
}
\subfigure[Outliers identified by RPC and WRPC.]{
\includegraphics[scale=0.35]{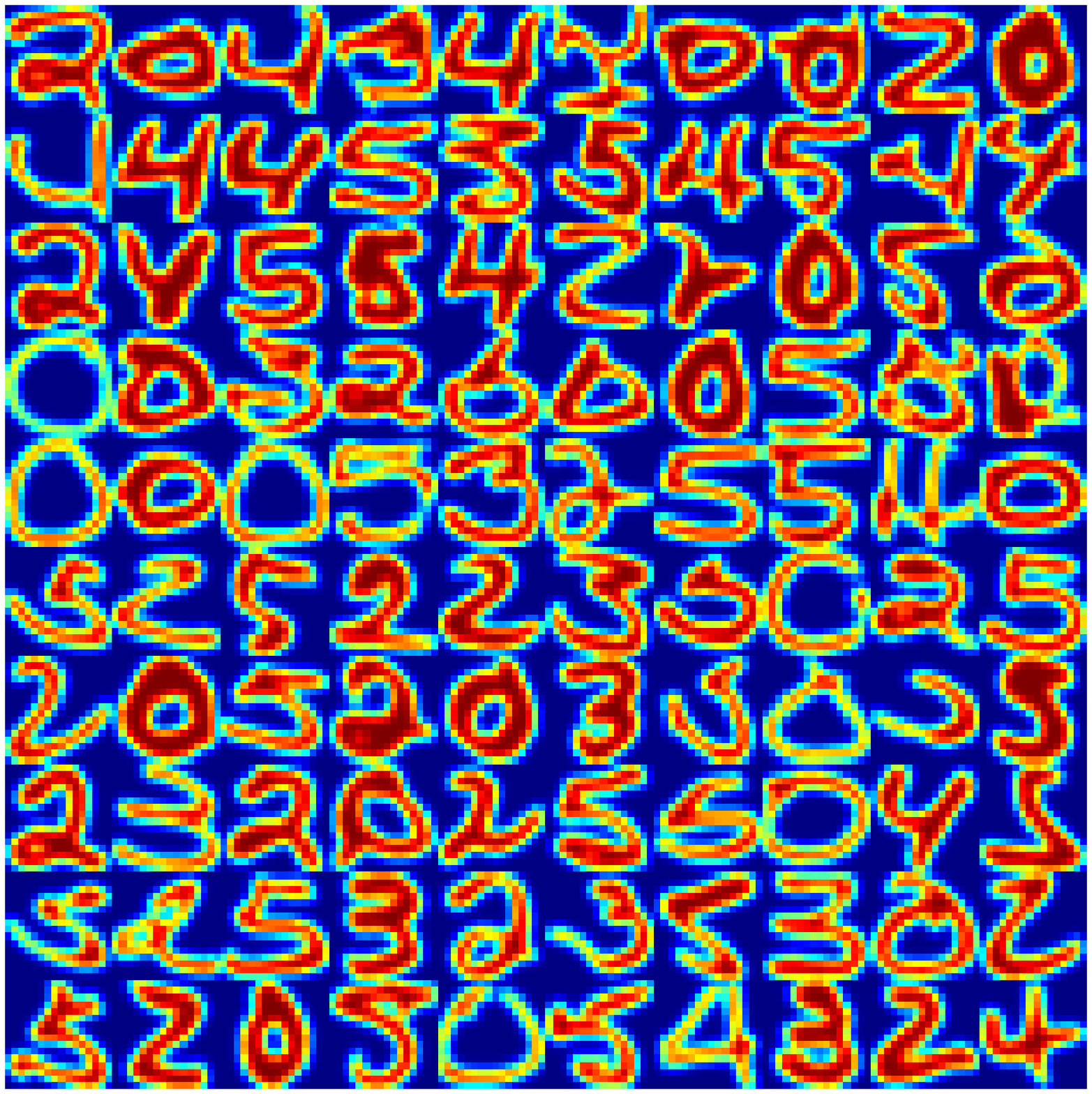}\label{fig:sim8_right}}
\subfigure[Outliers identified by KRKM using the polynomial kernel of order 3.]{
\includegraphics[scale=0.35]{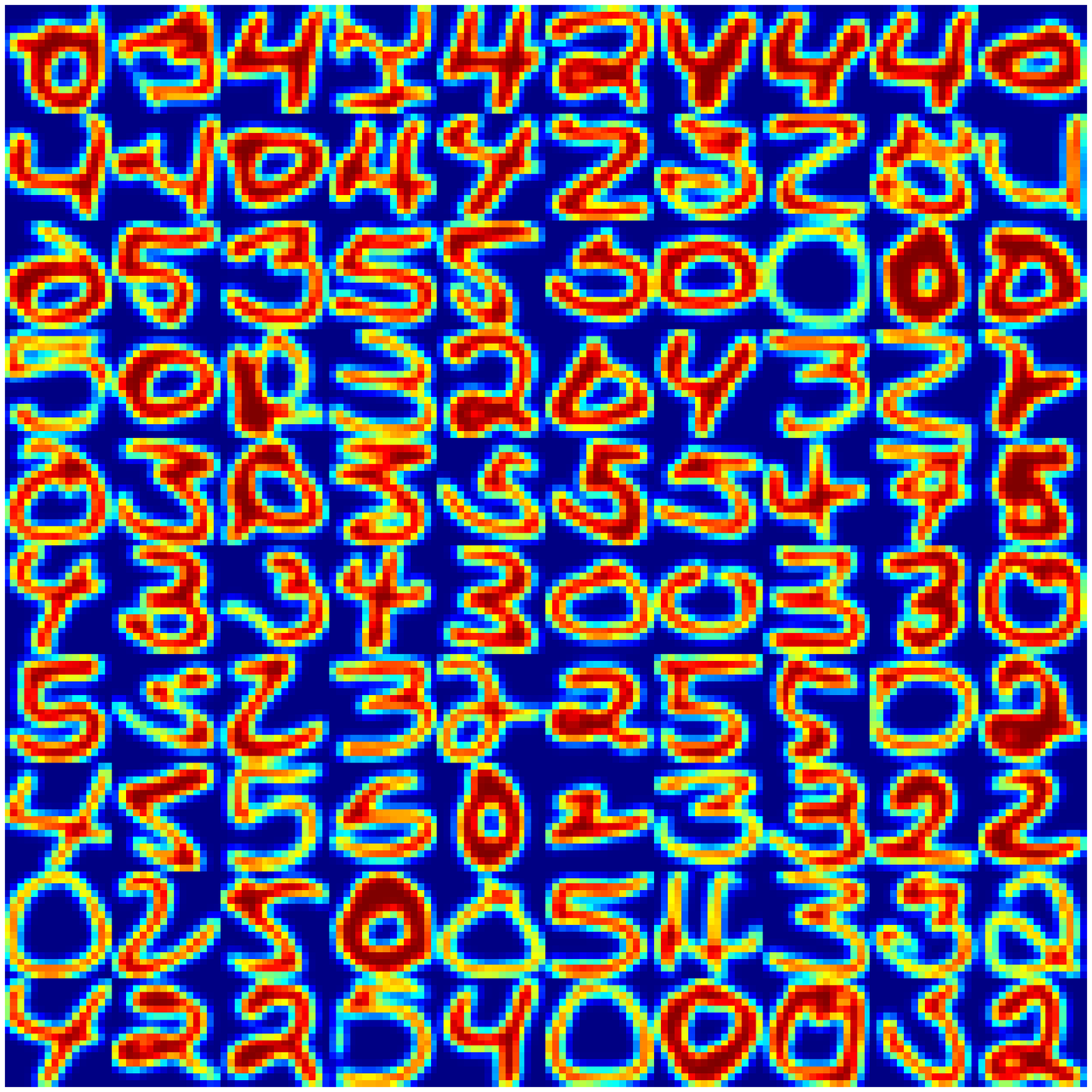}\label{fig:sim10}}

\caption{Clustering and outliers for the USPS dataset with $C=6$ tuned to identify $s=100$ outliers.}
\label{fig:sim8}
\end{figure}

The USPS dataset was clustered using the RKM and WRKM tuned to identify $100$ outliers. WRKM is initialized with the results obtained by RKM. Although RKM and WRKM yielded the same outlier images, the size of the $\obf_n$'s was different, becoming nearly uniform for WRKM. The USPS dataset was also clustered using the RPC and the WRPC algorithms. Fig.~\ref{fig:sim8_left} shows the cluster centroids obtained by RPC and WRPC. Fig.~\ref{fig:sim8_right} shows the $100$ outliers identified. The outliers identified by the RPC and WRPC algorithms also coincide. The position of the outlier images in the mosaic corresponds to their ranking according to the size of their corresponding $\obf_n$ (largest to smallest from left to right, top to bottom). Note that all outliers identified have a trait that differentiates them from the average image in each cluster.  Among the 100 outliers detected by RKM and RPC, 97 are common to both approaches.



A kernelized version of the algorithms was also used on the USPS dataset. Similar to \cite{ScSm02}, the homogeneous polynomial kernel of order 3, that is $\kappa(\xbf_n,\xbf_m)=(\xbf^T_n\xbf_m)^3$, was used. The ARI scores obtained by the kernelized robust clustering algorithms are shown in Table~\ref{table:t1}. Based on these scores, two important observations are in order: (i) kernelized K-means is more sensitive to outliers than K-means is; but (ii) KRKM for the particular kernel yields an improved clustering performance over RKM. Finally, the 100 outliers identified by KRKM are shown in Fig.~\ref{fig:sim10}.


\subsection{Dolphin's Social Network}\label{subsec:simdolph}
Next, KRKM is used to partition and identify outliers in a social network of $N=62$ bottlenose dolphins living in Doubtful Sound, New Zealand \cite{lune04dolphins}. Links between pairs of nodes (dolphins) represent statistically significant frequency association. The network structure is summarized by the $N\times N$ adjacency matrix $\mathbf{E}$. To identify social groups and outliers, the connection between kernel K-means and spectral clustering for graph partitioning is exploited \cite{Dhillon04kkmspecclust}. According to this connection, the conventional spectral clustering algorithm is substituted by the kernelized K-means algorithm with a specific kernel matrix. The kernel matrix used is $\Kbf=\nu\Ibf_N+\mathbf{D}^{-1/2} \mathbf{E}\mathbf{D}^{-1/2}$, where $\mathbf{D}:=\diag(\mathbf{E}\1bf_N)$ and $\nu$ is chosen larger than the minimum eigenvalue of $\mathbf{D}^{-1/2} \mathbf{E}\mathbf{D}^{-1/2}$ such that $\Kbf \succ 0$. 

\begin{figure}
\centering
\includegraphics[scale=0.6]{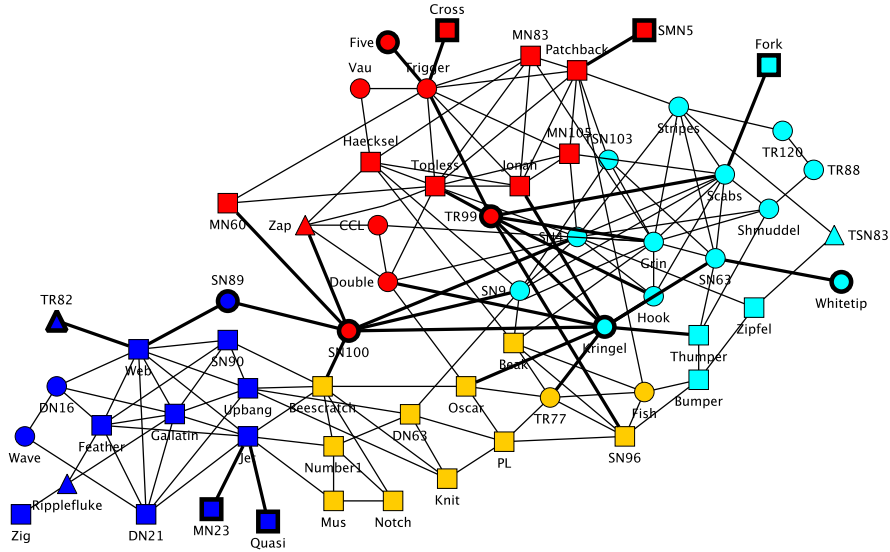}
\caption{KRKM clustering of the dolphin's social network: outliers are depicted bold-faced; male, female, and unobserved gender are represented by squares, circles, and triangles, respectively.}
\label{fig:dolphins1}
\end{figure}

Kernel K-means for graph partitioning is prone to being trapped at poor local minima, depending on initialization \cite{DhGuKu07}. The KRKM algorithm with $C=4$ clusters is initialized by the spectral clustering solution using the symmetric Laplacian matrix $\mathbf{L}:=\mathbf{I}_N- \mathbf{D}^{-1/2} \mathbf{E}\mathbf{D}^{-1/2}$. The parameter $\lambda$ is tuned to identify $s=12$ outliers. Fig.~\ref{fig:dolphins1} depicts the network partitioning and the outliers identified. The results show that several nodes identified as outliers have unit degree. Having a single link indicates that nodes are marginally connected to their corresponding clusters thus deemed as outliers. 

Other outlier instances adhere to more complicated structures within the network. Node \emph{zig} has a single link, yet it is not identified as an outlier possibly due to the reduced size of its cluster, especially since four other nodes in the same cluster are identified as outliers. Interestingly, nodes \emph{sn89}, \emph{sn100}, \emph{tr99} and \emph{kringel}, with node degrees 2, 7, 7, and 9, respectively, are also identified as outliers. Using gender information about the dolphins \cite{lune04dolphins}, we observe that \emph{sn89} is a female dolphin assigned to a cluster dominated by males (10 males, 3 females, and 2 unobserved). Likewise, the connectivity of \emph{sn100} and \emph{tr99} in the graph shows that they share many edges with female dolphins in other clusters which differentiates them from other female dolphins within the same cluster. Finally, \emph{kringel} is a dolphin connected to 6 dolphins in other clusters and only 3 dolphins in its own cluster.

\subsection{College Football Network}
\begin{figure}
\centering
\includegraphics[scale=0.25]{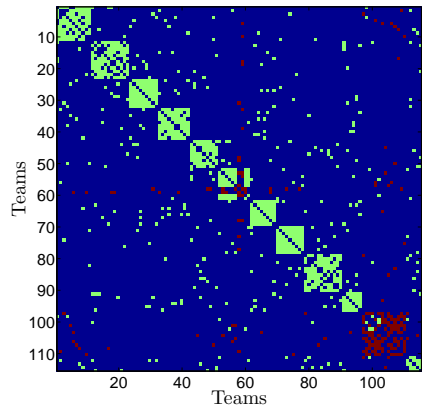}
\caption{The kernel matrix for the college football network permuted using KRKM clustering. Zero entries are colored blue and outliers are colored red.}
\label{fig:football1}
\end{figure}

KRKM is used to partition and identify outliers in a network of $N=115$ college football teams. The college football network represents the schedule of Division I games for the season in year 2,000 \cite{GirNew02}. Each node corresponds to a team and a link between two teams exists if they played against each other during the season. The teams are divided into $C=12$ conferences and each team plays games with teams in the same conference more often. KRKM is initialized via spectral clustering as described in Section \ref{subsec:simdolph}, while $\lambda$ is tuned to identify $s=12$ outliers. Fig.~\ref{fig:football1} shows the entries of the kernel matrix $\Kbf$ after being row and column permuted so that teams in the same cluster obtained by KRKM are consecutive. The ARI coefficient yielded by KRKM after removing the outliers was 0.9218. 

\begin{figure}
\centering
\includegraphics[scale=0.5]{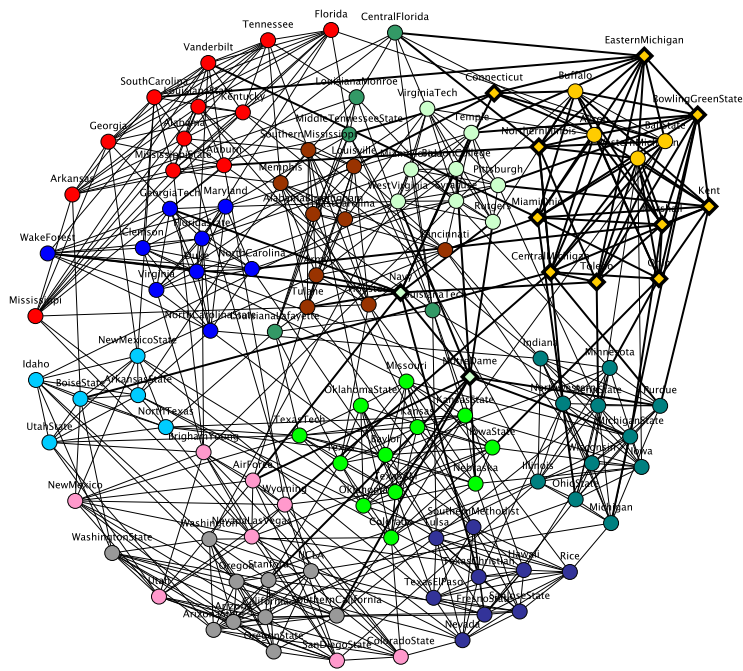}
\caption{Clustering of the college football network obtained by KRKM for $C=12$. Outliers are represented by diamond-shaped nodes.}
\label{fig:football2}
\end{figure}

Teams identified as outliers sorted in descending order based on their $\|\obf_n\|_2$ values are: Connecticut, Navy, Notre Dame, Northern Illinois, Toledo, Miami (Ohio), Bowling Green State, Central Michigan, Eastern Michigan, Kent, Ohio, and Marshall. Three of them, namely Connecticut, Notre Dame, and Navy, are independent teams. Connecticut is assigned to the Mid-American conference, but it does not play as many games with teams from this conference (4 games) as other teams in the same conference do (around 8 games). Notre Dame and Navy play an equal number of games with teams from two different conferences so they could be assigned to either one. Several teams from the Mid-American conference are categorized as outliers. In hindsight, this can be explained by the subdivision of the conference into East and West Mid-American conferences. Teams in each of the Mid-American sub-conferences played about the same number of games with teams from their own sub-conference and the rest of the teams. Interestingly, using KRKM with $C=13$ while still seeking for 12 outliers, the sub-partition of the Mid-American conference is identified. In this case, the ARI coefficient for the partition after removing outliers is 0.9110. The three independent teams, Connecticut, Notre Dame, and Navy, are again among the 12 outliers identified.

\section{Conclusions}\label{sec:conclusions}
Robust algorithms for clustering based on a principled data model accounting for outliers were developed. Both deterministic and probabilistic partitional clustering setups based on the K-means algorithm and GMM's, respectively, were considered. Exploiting the fact that outliers appear infrequently in the data, a neat connection with sparsity-aware signal processing algorithms was made. This led to the development of computationally efficient and provably convergent robust clustering algorithms. Kernelized versions of the algorithms, well-suited for high-dimensional data or when only similarity information among objects is available, were also developed. The performance of the robust clustering algorithms was validated via numerical experiments both on synthetic and real datasets.    

\bibliographystyle{IEEEtranS}
\bibliography{IEEEabrv,PAMIclustering}

\end{document}